\documentclass[11pt]{article}

\usepackage{acl}

\usepackage{times}
\usepackage{latexsym}

\usepackage[T1]{fontenc}

\usepackage[utf8]{inputenc}

\usepackage{microtype}

\usepackage{inconsolata}

\usepackage{graphicx}
\usepackage{booktabs}
\usepackage{times}
\usepackage{latexsym}
\usepackage{graphicx}
\usepackage{amsmath}
\usepackage{amssymb}
\usepackage{mathtools}
\usepackage{amsthm}
\usepackage{array}
\usepackage{longtable}
\usepackage[most]{tcolorbox}
\usepackage{bibentry}
\usepackage{bm} 
\usepackage{tabularx}
\usepackage{multirow} 
\usepackage{algorithm}
\usepackage{algorithmic}
\usepackage{arydshln}
\usepackage{appendix}
\usepackage{wrapfig}   
\usepackage{bbm}
\usepackage{xcolor}
\usepackage{enumitem}
\usepackage[dvipsnames]{xcolor}
\usepackage{tcolorbox}
\tcbuselibrary{listings}
\tcbuselibrary{minted}
\usepackage[T1]{fontenc}
\makeatletter
\def\blankfootnote{\xdef\@thefnmark{}\@footnotetext}
\makeatother

\definecolor{royalblue(web)}{rgb}{0.25, 0.41, 0.88}
\definecolor{blue-violet}{rgb}{0.54, 0.17, 0.89}
\definecolor{brightmaroon}{rgb}{0.76, 0.13, 0.28}
\definecolor{darkmagenta}{rgb}{0.55, 0.0, 0.55}
\definecolor{bleudefrance}{rgb}{0.19, 0.55, 0.91}
\definecolor{palatinateblue}{rgb}{0.15, 0.23, 0.89}
\definecolor{royalblue(web)}{rgb}{0.25, 0.41, 0.88}
\definecolor{whitesmoke}{rgb}{0.96, 0.96, 0.96}
\definecolor{thulianpink}{rgb}{0.87, 0.44, 0.63}
\definecolor{amber(sae/ece)}{rgb}{1.0, 0.49, 0.0}
\definecolor{darkblue}{rgb}{0.0, 0.0, 0.55}
\definecolor{alizarin}{rgb}{0.82, 0.1, 0.26}
\definecolor{asparagus}{rgb}{0.53, 0.66, 0.42}
\definecolor{darkspringgreen}{rgb}{0.09, 0.45, 0.27}
\definecolor{columbiablue}{rgb}{0.61, 0.87, 1.0}
\definecolor{wildblueyonder}{rgb}{0.64, 0.68, 0.82}
\definecolor{trolleygrey}{rgb}{0.5, 0.5, 0.5}
\definecolor{paleaqua}{rgb}{0.74, 0.83, 0.9}
\definecolor{bubblegum}{rgb}{0.99, 0.76, 0.8}
\definecolor{coralred}{rgb}{1.0, 0.25, 0.25}
\definecolor{green(ryb)}{rgb}{0.4, 0.69, 0.2}
\definecolor{flame}{rgb}{0.89, 0.35, 0.13}
\definecolor{bittersweet}{rgb}{1.0, 0.44, 0.37}
\definecolor{darksalmon}{rgb}{0.91, 0.59, 0.48}
\definecolor{emerald}{rgb}{0.31, 0.78, 0.47}
\definecolor{green(pigment)}{rgb}{0.0, 0.65, 0.31}
\definecolor{codegreen}{rgb}{0,0.6,0}
\definecolor{codegray}{rgb}{0.5,0.5,0.5}
\definecolor{codepurple}{rgb}{0.58,0,0.82}
\definecolor{backcolour}{rgb}{0.96,0.96,0.94}
\definecolor{bluegray}{rgb}{0.3, 0.38, 0.47}
\definecolor{whitesmoke}{rgb}{0.96, 0.96, 0.96}
\definecolor{codegreen}{rgb}{0,0.6,0}
\definecolor{codegray}{rgb}{0.5,0.5,0.5}
\definecolor{codepurple}{rgb}{0.58,0,0.82}
\definecolor{backcolour}{rgb}{0.96,0.96,0.94}
\definecolor{darkred}{rgb}{0.8, 0, 0}

\tcbuselibrary{breakable}
\usepackage{listings}
\lstdefinestyle{mystyle}{
  basicstyle=\scriptsize\ttfamily,
  frame=single, 
  columns=fixed, 
}

\lstdefinestyle{newstyle}{
  basicstyle=\footnotesize\ttfamily\color{codegreen},
  backgroundcolor=\color{backcolour},
  frame=shadowbox, 
  rulecolor=\color{red},
  frameround=tttt, 
  keywordstyle=\color{magenta},
  commentstyle=\color{green},
  stringstyle=\color{red},
  showstringspaces=false,
  numbers=left,
  numberstyle=\tiny\color{gray},
  breaklines=true
}

\usepackage{color}

\newtheorem{theorem}{Theorem}

\newcommand{\ours}{{\fontfamily{qpl}\selectfont LFPO}}
\DeclareTextFontCommand{\textbf}{\fontfamily{qpl}\bfseries\selectfont}
\DeclareTextFontCommand{\textit}{\fontfamily{qpl}\itshape\selectfont}

\renewcommand{\eqref}[1]{(\ref{#1})}

\usepackage{amsmath,amsfonts,bm}

\def\1{\bm{1}}

\DeclareMathAlphabet{\mathsfit}{\encodingdefault}{\sfdefault}{m}{sl}
\SetMathAlphabet{\mathsfit}{bold}{\encodingdefault}{\sfdefault}{bx}{n}

\title{Efficient Token-Level Policy Optimization for Masked Diffusion Models via Velocity Alignment}

\author{
\bfseries
Chenxing Wei$^{\dagger,1,2,3}$ \quad
Jiazhen Kang$^{4,3}$ \quad
Hong Wang$^{5}$ \quad
Jianqing Zhang$^{6}$ \quad
Hao Jiang$^{5,3}$ \\
\textbf{Xiaolong Xu}$^{3}$ \quad
\textbf{Ningyuan Sun}$^{3}$ \quad
\textbf{Ying He}$^{8}$ \quad
\textbf{Fei Yu}$^{7}$ \quad
\textbf{Yao Shu}$^{*,9}$ \quad
\textbf{Bo Jiang}$^{*,3}$ \\
$^1$The University of Hong Kong\quad
$^2$Shenzhen Loop Area Institute \quad
$^3$ByteDance \\
$^4$Southeast University \quad
$^5$University of Science and Technology of China \\
$^6$Shanghai Jiao Tong University \quad
$^7$Carleton University \quad 
$^8$Shenzhen University \quad \\
$^9$The Hong Kong University of Science and Technology (Guangzhou) \\
$^*$Corresponding authors, $\dagger$ Work done during an internship at ByteDance. \\
\texttt{weichenxing@connect.hku.hk}, \texttt{yaoshu@hkust-gz.edu.cn}\\
\texttt{jiangbo.jacob@bytedance.com}
}

\begin{document}
\maketitle
\begin{abstract}
Reinforcement Learning with Verifiable Rewards (RLVR) successfully aligns autoregressive models in rigorous domains like mathematical reasoning and code generation. However, applying RLVR to Diffusion Large Language Models (dLLMs) is hindered by intractable sequence-level likelihoods, forcing existing methods to rely on unstable, high-variance approximations. We propose \textit{Sequence \underline{L}ikelihood-\underline{F}ree \underline{P}olicy \underline{O}ptimization} (\ours{}), an efficient framework that circumvents this bottleneck by formulating policy alignment as \textit{discrete velocity alignment}. Inspired by continuous flow matching, \ours{} eschews trajectory-level estimation and directly optimizes denoising logits via contrastive token-level updates, providing a stable, low-variance surrogate. Furthermore, predicting terminal solutions from intermediate states implicitly straightens the generation trajectory, enabling high-quality sampling with significantly fewer denoising iterations. Extensive experiments on reasoning and coding benchmarks demonstrate that \ours{} matches or exceeds the accuracy of likelihood-based baselines while reducing denoising steps and improving practical decoding throughput.
\end{abstract}

\section{Introduction}
\label{sec:introduction}

\begin{figure*}[t]
\vspace{-1mm}
\centering
\includegraphics[width=1.0\textwidth, trim={1cm 5.5cm 1cm 5cm}, clip]{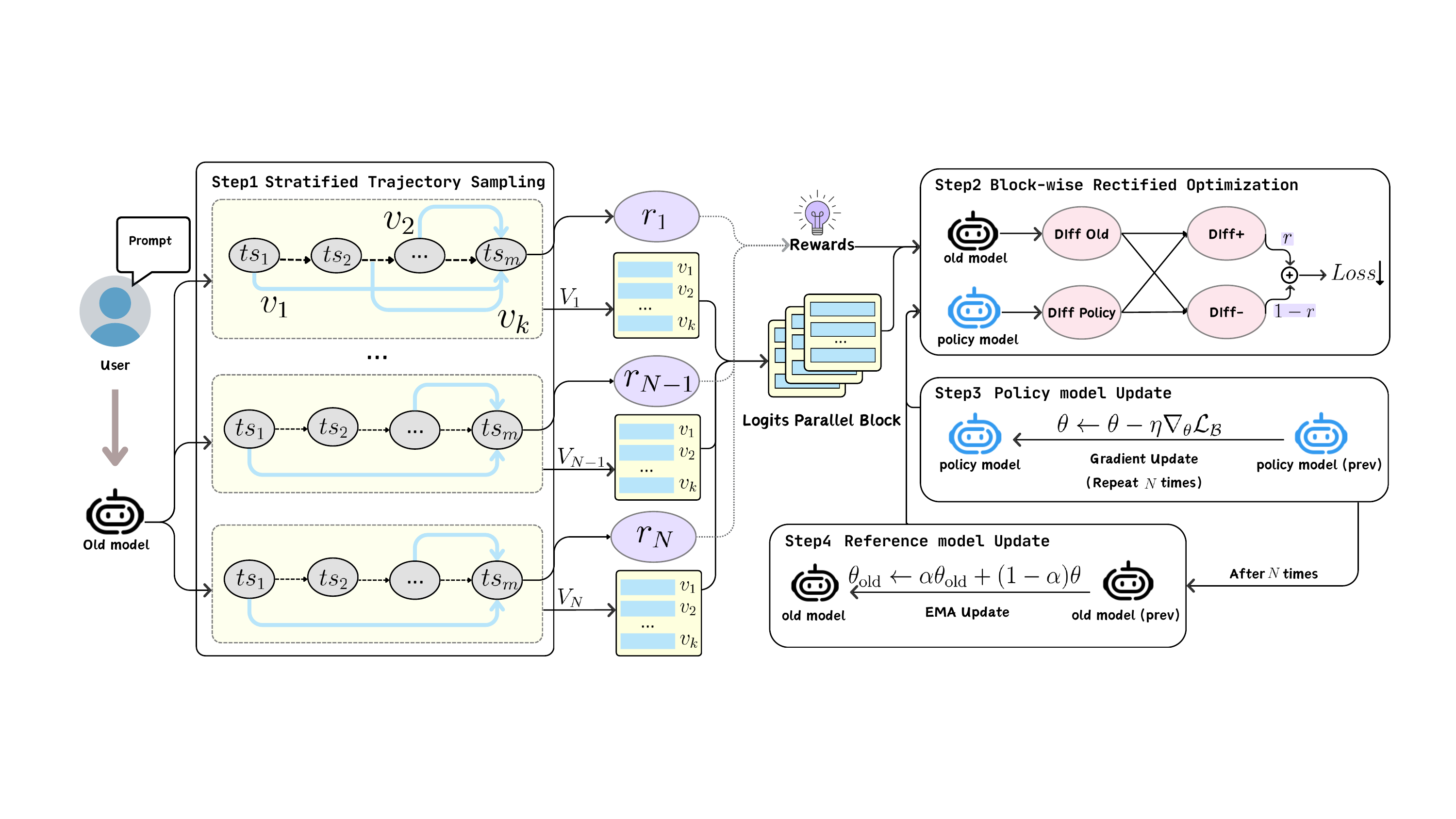}
\vspace{-3mm}
\caption{\textbf{Overview of the \ours{} framework.} \textbf{Step 1} \textit{Generate \& Estimate}: $\pi_{\text{old}}$ generates trajectories, with timesteps selected via \textit{Stratified Trajectory Sampling} to minimize variance. \textbf{Step 2} \textit{Block-Wise Optimization}: Data is partitioned into memory-efficient blocks for parallel logit computation. \textbf{Step 3} \textit{Policy Update}: $\pi_\theta$ minimizes deviation from reward-induced implicit policies ($\pi^+, \pi^-$) via discrete vector field rectification. \textbf{Step 4} \textit{Reference Update}: Stable EMA update of $\pi_{\text{old}}$. See Appendix~\ref{app:method_details} for method and Appendix~\ref{app:hyperparams} for implementation details.}
\label{fig:method}
\vspace{-3mm}
\end{figure*}

While \textit{autoregressive} (AR) models have long dominated mathematical reasoning and code generation~\cite{hui2024qwen2, Gemini, Claude}, \textit{Diffusion Large Language Models} (dLLMs)~\cite{ye2025dream, nie2025large} have recently emerged as a compelling alternative~\cite{gong2025diffucoderunderstandingimprovingmasked}. Fundamentally diverging from the strict left-to-right sequential generation~\cite{tian2024visual}, dLLMs operate via a holistic denoising mechanism~\cite{li2025surveydiffusionlanguagemodels}. This non-autoregressive paradigm endows dLLMs with inherent advantages in global planning~\cite{ye2025beyond} and iterative refinement, enabling simultaneous updates across the entire sequence structure~\cite{havasi2025edit, zhang2023planner}. Furthermore, dLLMs facilitate significantly faster inference through parallel decoding strategies~\cite{wu2025fastdllmtrainingfreeaccelerationdiffusion, wang2025diffusionllmsfasterthanarinference}. Despite these capabilities, effectively aligning dLLMs with verifiable correctness feedback remains a formidable open challenge~\cite{zhao2025d, zhan2025principledtractablerlreasoning}.

\textit{Reinforcement Learning with Verifiable Rewards} (RLVR)~\cite{shao2024deepseekmathpushinglimitsmathematical} is the gold standard for aligning AR models in reasoning tasks~\cite{wei2025redit, zhang2026gaporobustadvantageestimation}. Consequently, recent methods attempt to transpose this paradigm to dLLMs~\cite{zhao2025d, zhan2025principledtractablerlreasoning,ni2026flexibilitytrap,ou2025principledsequence,rojas2025improvingreasoninggdpo} by casting the non-Markovian diffusion process as a \textit{Markov Decision Process} (MDP)~\cite{yang2025mmada, wang2025revolutionizingreinforcementlearningframework} to apply \textit{Policy Gradient} (PG) algorithms~\cite{schulman2017proximalpolicyoptimizationalgorithms}. However, standard PG estimation relies on exact model likelihoods to compute importance sampling weights~\cite{schulman2017trustregionpolicyoptimization, zheng2025groupsequencepolicyoptimization}, posing a critical bottleneck due to the computationally intractable sequence-level log-likelihoods of dLLMs~\cite{wang2025spgsandwichedpolicygradient, zhao2025d}. Lacking exact trajectory likelihoods, current approaches resort to step-wise ODE/SDE approximations~\cite{chen2023the, song2021scorebasedgenerativemodelingstochastic, gong2025diffucoderunderstandingimprovingmasked}. In the high-dimensional discrete token space, these approximations inevitably introduce severe error accumulation~\cite{zhao2025d} and high computational overhead~\cite{wang2025revolutionizingreinforcementlearningframework}, resulting in training instability and sub-optimal efficiency~\cite{ni2025trainingoptimallargediffusion} (see Appendix~\ref{sec:related_work} for extended related work).

Unlike autoregressive models, whose sequential inference naturally fits the MDP paradigm for straightforward likelihood computation~\cite{xiong2025deepseekparadigmshiftstechnical}, imposing trajectory-level likelihood estimation on dLLMs is inherently inefficient, as their holistic denoising mechanism renders exact evaluation intractable~\cite{li2024likelihood, zhu2025llada15variancereducedpreference}. To circumvent this bottleneck, we analyze dLLMs through the geometric lens of \textit{Flow Matching} (FM)~\cite{lipman2023flow}, establishing that predicted logits for masked tokens effectively serve as a discrete proxy for the continuous velocity field $v$~\cite{loudiscretediffusion}. Consequently, rather than approximating the intractable sequence integral $P_\theta(x)$, we propose aligning policies by directly rectifying these discrete logits toward high-reward trajectories via token-level \textit{velocity alignment}. By performing localized updates in the logit space, we bypass the reliance on trajectory-level probability estimation, a strategy that aligns with recent breakthroughs in visual diffusion reinforcement learning~\cite{tuo2025scalablemultitemperaturefreeenergy, zheng2026diffusionnft}.

To actualize this vision, we propose \textbf{Sequence Likelihood-Free Policy Optimization (\ours{})}, an efficient token-level framework for aligning masked diffusion models. As illustrated in Figure~\ref{fig:method}, \ours{} circumvents the sequence-likelihood bottleneck by operating directly on denoising logits. Employing a contrastive objective, it rectifies the denoising vector field by pulling predictions toward high-reward targets while repelling them from suboptimal ones. By eschewing noisy trajectory likelihood approximations, \ours{} establishes a localized, low-variance empirical surrogate that significantly stabilizes training. Furthermore, to mitigate the inherent instability of step-wise denoising, \ours{} incorporates a terminal anchoring mechanism that maps intermediate states directly to the final solution. This consistency objective encourages straighter probability flows, suppressing trajectory fluctuations and enabling high-quality sampling with drastically fewer iterations. Empirically, \ours{} yields substantial accuracy improvements across reasoning and coding tasks while reducing denoising steps and improving practical throughput. Our core contributions are threefold:
\begin{itemize}
    \item \textbf{Geometric Perspective:} Inspired by FM, we cast denoising logits as a discrete velocity proxy, deriving a token-level surrogate objective that entirely avoids intractable trajectory likelihood integration (Section~\ref{sec:motivation}).
    \item \textbf{\ours{} Framework:} We propose a highly efficient RL framework that optimizes dLLMs via contrastive velocity alignment. This localized approach bypasses complex ODE backtracking, enabling robust and scalable training (Section~\ref{sec:method}; see Appendix~\ref{app:method_details} for supplementary conceptual discussions).
    \item \textbf{Empirical Superiority:} \ours{} achieves a superior accuracy-efficiency trade-off on reasoning and coding benchmarks, matching or exceeding likelihood-based baselines while reducing denoising steps and improving examples/s and generated tokens/s (Section~\ref{sec:results}; Appendix~\ref{app:inference_hyperparams}).
\end{itemize}

\section{Motivation: A Flow Matching Perspective}
\label{sec:motivation}

Standard reinforcement learning approaches for diffusion models are hindered by the intractability of likelihood computation. To overcome this, we reframe policy alignment through the geometric lens of FM. This section first reviews continuous FM (Section~\ref{subsec:continuous_fm}), lifts discrete tokens to a probability simplex (Section~\ref{subsec:discrete_lifting}), and presents a pointwise logit-gradient identity that motivates our token-level surrogate (Section~\ref{subsec:equivalence}).

\subsection{Flow Matching}
\label{subsec:continuous_fm}

FM provides a simulation-free paradigm for modeling a probability path from a simple prior $p_0$ to a complex data distribution $p_1$ via an Ordinary Differential Equation (ODE): $\frac{d\boldsymbol{x}}{dt} = v_t(\boldsymbol{x})$ with $\boldsymbol{x}(0) \sim p_0$. The goal is to regress a neural vector field $v_\theta(\boldsymbol{x}_t, t)$ to match a target conditional vector field $u_t$. Under the standard \textit{Conditional Optimal Transport} path~\cite{onken2021otflowfastaccuratecontinuous}, which interpolates linearly as $\boldsymbol{x}_t = (1-t)\boldsymbol{x}_0 + t\boldsymbol{x}_1$, the target velocity becomes a constant displacement:
\begin{equation}
    u_t(\boldsymbol{x}_t | \boldsymbol{x}_1) = \frac{d\boldsymbol{x}_t}{dt} = \boldsymbol{x}_1 - \boldsymbol{x}_0.
\end{equation}
We minimize the expected \textit{Mean Squared Error} between these velocities:
\begin{equation}
    \mathcal{L}_{FM}(\theta) = \mathbb{E}_{t, \boldsymbol{x}_1, \boldsymbol{x}_0} \left[ \| v_\theta(\boldsymbol{x}_t, t) - (\boldsymbol{x}_1 - \boldsymbol{x}_0) \|^2 \right].
\end{equation}

\subsection{Lifting Discrete Tokens to the Probability Simplex}
\label{subsec:discrete_lifting}

\begin{figure}[t]
\vspace{-1mm}
\centering
\includegraphics[width=0.48\textwidth]{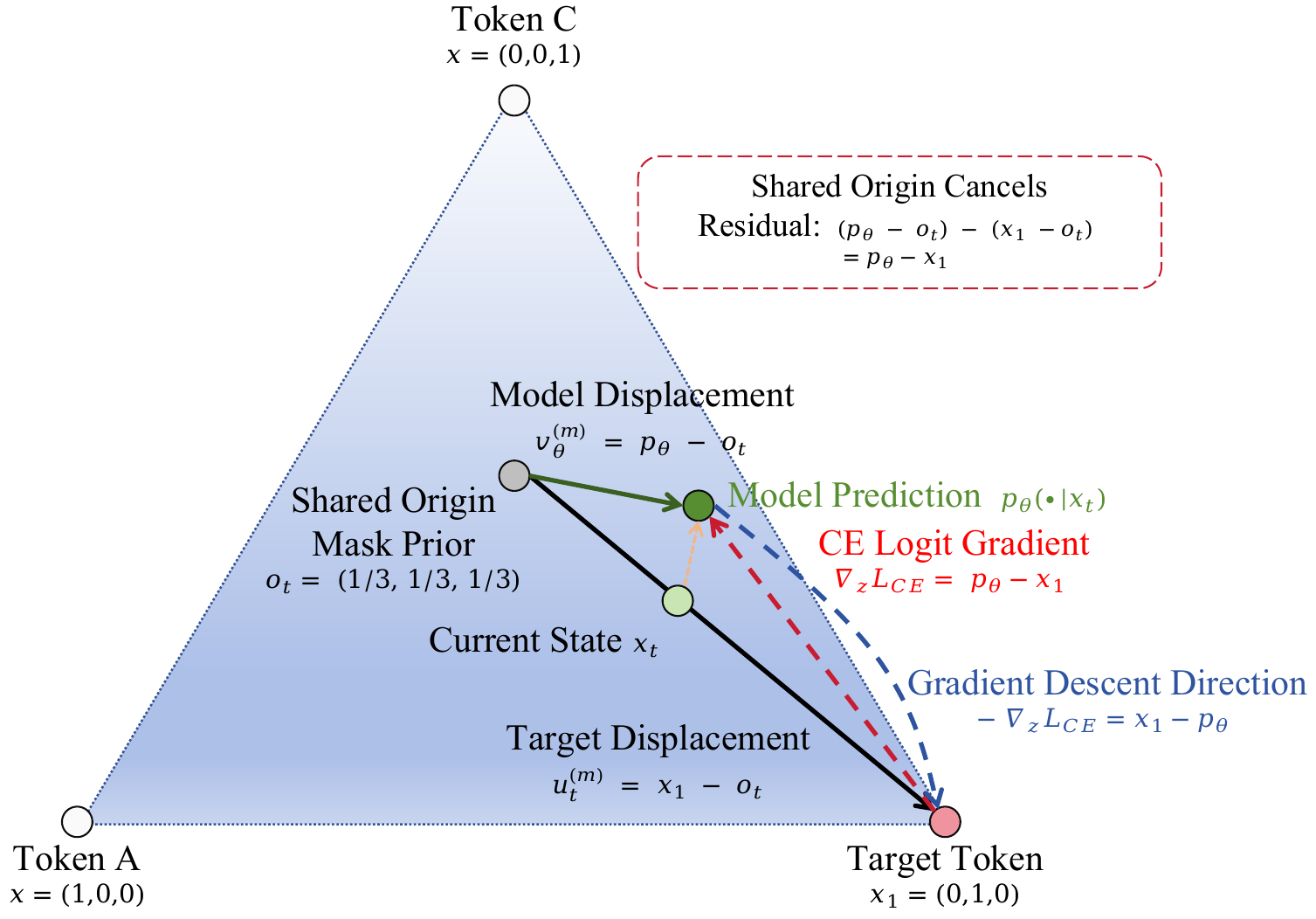}
\vspace{-2mm}
\caption{\textbf{Geometric Perspective of Discrete Lifting.} Visualization on a 2-simplex ($\Delta^{2}$) where vertices represent tokens. Using a shared origin $\boldsymbol{o}_t$, the \textit{Target Velocity} $u_t^{(o)}$ (black) points toward the ground truth $\boldsymbol{x}_1$, while the \textit{Model Velocity} $v_\theta^{(o)}$ (green) points toward the model prediction $p_\theta$. The dashed line indicates the origin-invariant residual direction, $p_\theta-\boldsymbol{x}_1$, which corresponds exactly to the CE logit gradient $\nabla \mathcal{L}_{CE}$.}
\label{fig:probability_simplex}
\vspace{-3mm}
\end{figure}

To bridge the gap between continuous FM and discrete dLLMs, we embed the vocabulary $\mathcal{V}$ as vertices on a probability simplex $\Delta^{V-1} \subset \mathbb{R}^V$ (Figure~\ref{fig:probability_simplex}). In this geometric formulation, data tokens are represented as deterministic one-hot vectors $\boldsymbol{x}_1 \in \{0, 1\}^V$ at the vertices, while the \texttt{[MASK]} token is defined as a fixed prior $\boldsymbol{m} \in \Delta^{V-1}$ at the simplex center (typically a uniform distribution). The forward diffusion process—transitioning from a masked to a revealed state—is modeled as a linear interpolation trajectory: $\boldsymbol{x}_t = (1 - \alpha_t) \boldsymbol{m} + \alpha_t \boldsymbol{x}_1$, where $\alpha_t \in [0, 1]$ dictates the noise schedule. 

Because our token-level objective relies solely on the residual between the model and target directions, we define both displacements relative to an arbitrary shared origin $\boldsymbol{o}_t$ on the simplex (e.g., the mask prior $\boldsymbol{m}$ or the current state $\boldsymbol{x}_t$). This ensures the residual remains origin-invariant. The target displacement is thus defined as:
\begin{equation}
    u_t^{(o)} \coloneqq \boldsymbol{x}_1 - \boldsymbol{o}_t.
\end{equation}

Crucially, a dLLM outputs a probability distribution $p_\theta(\cdot | \boldsymbol{x}_t) = \text{Softmax}(\text{Logits}_t)$ over the vocabulary (depicted as the point $p_\theta$ in the simplex). Analogous to the continuous case, we define the \textbf{model velocity field} $v_\theta^{(o)}$ as the vector displacement from the shared origin to the prediction:
\begin{equation}
    v_\theta^{(o)}(\boldsymbol{x}_t, t) \coloneqq p_\theta(\cdot | \boldsymbol{x}_t) - \boldsymbol{o}_t.
\end{equation}
This formulation interprets dLLM logits as local displacement parameters, rendering the velocity residual inherently origin-independent.

\subsection{Pointwise Logit-Gradient Identity}
\label{subsec:equivalence}

We use a pointwise identity linking standard token-level \textit{Cross-Entropy} (CE) training to this geometric perspective. By evaluating the expected squared error between the model and target velocities, the shared origin $\boldsymbol{o}_t$ naturally cancels out:

\begin{equation}
\begin{split}
    \mathcal{L}_{\mathrm{res}} & = \mathbb{E} \left[ \| (p_\theta - \boldsymbol{o}_t) - (\boldsymbol{x}_1 - \boldsymbol{o}_t) \|^2 \right]  \\ &= \mathbb{E} \left[ \| p_\theta - \boldsymbol{x}_1 \|^2 \right].
\end{split}
\label{eq:velocity_residual}
\end{equation}

Standard dLLM training minimizes the token-level CE loss:
\begin{equation}
\begin{split}
    \mathcal{L}_{CE} &= \mathbb{E} \left[ - \sum_{i} (\boldsymbol{x}_1)_i \log (p_\theta)_i \right] \\ &= \mathbb{E} \left[ - \log (p_\theta)_k \right],
\end{split}
\label{eq:ce_loss}
\end{equation}
where $k$ is the index of the ground-truth token. While these objectives reside in distinct functional spaces (log-likelihood versus $L_2$ distance) and are not globally equivalent, their pointwise logit gradients are mathematically identical:

\begin{tcolorbox}[colback=royalblue(web)!3!white,colframe=royalblue(web)!90!white, left=0.5mm, right=1mm, top=1mm, bottom=1mm]
\begin{theorem}[Pointwise Logit-Gradient Identity]
\label{thm:gradient_alignment}
Let $\boldsymbol{z} \in \mathbb{R}^V$ be the unconstrained pre-softmax logits such that $p_\theta = \text{Softmax}(\boldsymbol{z})$. For any discrete time step $t$ and any shared origin $\boldsymbol{o}_t$ fixed for the current input state, the gradient of $\mathcal{L}_{CE}$ with respect to the logits $\boldsymbol{z}$ exactly matches the residual error vector between the model velocity $v_\theta^{(o)}$ and the target velocity $u_t^{(o)}$:
\begin{equation}
    \nabla_{\boldsymbol{z}} \mathcal{L}_{CE} = p_\theta - \boldsymbol{x}_1 = v_\theta^{(o)} - u_t^{(o)}.
\end{equation}
\end{theorem}
\end{tcolorbox}

\noindent\textbf{Remark.} The detailed proof is provided in Appendix~\ref{app:gradient_derivation} and Appendix~\ref{app:theory_scope} details its theoretical scope and bounds the logit gradient. Theorem~\ref{thm:gradient_alignment} establishes a pointwise logit-gradient identity rather than global equivalence between the CE and FM objectives. Thus, our contribution does not rely on an exact FM--CE objective equivalence; instead, we use the identity as a local design principle for constructing a tractable token-level RLVR surrogate. To bridge the distributional gap between the discrete masking process and the continuous FM interpolation path, we sample intermediate states directly from the model's empirical generation trajectories. Motivated by this empirical approximation, \ours{} (Section~\ref{sec:method}) employs a contrastive objective to rectify the velocity field by pulling the model velocity $v_\theta^{(o)}$ toward high-reward Monte Carlo samples while repelling it from low-reward ones.

\section{Policy Alignment via Velocity Alignment}
\label{sec:method}

We present \ours{}, an RL framework for dLLMs. While inspired by continuous FM, its direct application is bottlenecked by the intractability of continuous ODEs in discrete token spaces. The pointwise logit-gradient identity in Theorem~\ref{thm:gradient_alignment} motivates an efficient implementation of velocity alignment via contrastive token-level updates.

\subsection{Contrastive Velocity Alignment}
\label{subsec:implicit_policy}

In the supervised setting (Section~\ref{sec:motivation}), the optimization target is explicit: the ground-truth data defines a clear target displacement under the shared-origin convention (e.g., $\boldsymbol{u}_t^{(o)} = \boldsymbol{x}_1 - \boldsymbol{o}_t$). However, in the RLVR setting, such ground truth is absent. The model interacts with the environment to sample a trajectory $\tau$, receiving only a sparse scalar reward $r(\tau)$. The fundamental challenge thus lies in defining a valid supervision target to effectively identify the correct velocity direction based solely on this scalar feedback. To address this, we draw inspiration from~\cite{zheng2026diffusionnft}, which constructs implicit target velocity fields to guide continuous flow matching. We adapt this formulation into the discrete logit space. A critical design choice in \ours{} is performing policy perturbations in the logit space rather than the probability space.

\noindent\textbf{Justification for Log-Space Perturbation.} As summarized in Table~\ref{tab:perturbation_comparison}, applying linear perturbations directly in the probability space (e.g., $\pi' = \text{clip}(\pi_{\text{ref}} + \beta \Delta, 0, 1)$) frequently violates the $[0, 1]$ boundary constraints. This necessitates heuristic truncation, which fragments gradient continuity and yields an unstable optimization landscape. Conversely, performing perturbations in the logit space naturally bounds the policy within the simplex via the Softmax operation, maintaining a smooth and differentiable optimization landscape (see the expanded discussion on simplex constraints in Appendix~\ref{app:method_details}).

Let $\pi_{\text{ref}}$ denote a frozen reference policy and $\pi_\theta$ the active policy. We define the velocity deviation as $\Delta(\boldsymbol{x}_t) = \log \pi_\theta(\cdot|\boldsymbol{x}_t) - \log \pi_{\text{ref}}(\cdot|\boldsymbol{x}_t)$. Based on this, we explicitly construct two implicit target policies in the log-space:
\begin{align}
    \log \pi^+(\cdot|\boldsymbol{x}_t) &\coloneqq \log \pi_{\text{ref}}(\cdot|\boldsymbol{x}_t) + \beta \Delta(\boldsymbol{x}_t), \label{eq:pos_policy} \\
    \log \pi^-(\cdot|\boldsymbol{x}_t) &\coloneqq \log \pi_{\text{ref}}(\cdot|\boldsymbol{x}_t) - \beta \Delta(\boldsymbol{x}_t), \label{eq:neg_policy}
\end{align}
where $\beta > 0$ is a scalar coefficient. Geometrically, $\pi^+$ (\textit{Implicit Positive Policy}) amplifies the current model's deviation from the reference, whereas $\pi^-$ (\textit{Implicit Negative Policy}) reverses it. These detached definitions provide the necessary gradient targets for the RLVR setting. For a detailed conceptual positioning against related diffusion RL and FM frameworks (Table~\ref{tab:contribution_decomposition}), as well as a geometric comparison with naive reward-weighted cross-entropy (Table~\ref{tab:fm_vs_ce}), please refer to Appendix~\ref{app:method_details}.

Leveraging the scalar reward $r(\tau) \in [0, 1]$, we formulate the \ours{} objective as a dynamic interpolation between these implicit targets. We minimize the reward-weighted CE loss between the model's velocity field and these targets:
\begin{equation}
\begin{split}
    \mathcal{L}_{\text{\ours{}}}(\theta) = & \mathbb{E}_{\tau \sim \pi_{\theta}} \Big[ \mathbb{E}_{t} \Big[  r(\tau) \cdot \text{CE}(\text{sg}(\pi^+), \pi_\theta) \\
    & + (1 - r(\tau)) \cdot \text{CE}(\text{sg}(\pi^-), \pi_\theta) \Big] \Big],
\end{split}
\label{eq:lfpo_loss}
\end{equation}
where $\text{CE}(P, Q) = - \sum P(x) \log Q(x)$ and $\text{sg}(\cdot)$ denotes the stop-gradient operation. Crucially, because $\pi^+$ and $\pi^-$ depend on the active policy $\pi_\theta$, these implicit target policies are computed without backpropagating through the target construction, ensuring unambiguous gradients and stable velocity rectification. Intuitively, high rewards ($r \to 1$) drive the policy to align with $\pi^+$ (pull), while low rewards ($r \to 0$) drive it toward $\pi^-$ (push). 

\paragraph{Reward-bearing trajectories.}
The scalar reward enters token-level supervision through sampled terminal completions rather than through an assumed per-token reward. For each prompt $q$, we sample rollouts from the behavior policy and evaluate them with the verifier:
\begin{equation}
    y^{(j)} \sim \pi_{\mathrm{old}}(\cdot \mid q), \qquad
    r^{(j)} = R(q,y^{(j)}) \in [0,1].
\end{equation}
We then reuse the intermediate masked states $\boldsymbol{x}_t^{(j)}$ along each reward-bearing trajectory to construct local logit-space targets:
\begin{equation}
\begin{split}
    \Delta_t^{(j)} = \log \pi_\theta(\cdot \mid &\boldsymbol{x}_t^{(j)},q)
    - \log \pi_{\mathrm{ref}}(\cdot \mid \boldsymbol{x}_t^{(j)},q), \\
    \log \pi_{j,t}^{+} &= \log \pi_{\mathrm{ref}} + \beta \Delta_t^{(j)}, \\
    \log \pi_{j,t}^{-} &= \log \pi_{\mathrm{ref}} - \beta \Delta_t^{(j)}.
\end{split}
\end{equation}
Thus, successful terminal completions provide pull targets at their visited token states, while failed completions provide push targets at their visited token states. At the exact degenerate point where the active and reference policies are identical and all sampled completions have identical rewards, \ours{} has no meaningful preference signal; in practice, reward diversity from verifiable rollouts and the lagged EMA reference break this symmetry after early updates. We make this cold-start dependence explicit in the limitations.

Directly optimizing Eq.~(\ref{eq:lfpo_loss}) presents practical challenges. Naively sampling a single timestep $t$ yields high gradient variance and optimization instability. To mitigate this, we employ \textit{Stratified Trajectory Sampling} to ensure dense temporal coverage. However, this multi-sample approach imposes a severe GPU memory burden. Consequently, we introduce \textit{Block-wise Gradient Accumulation} to resolve this memory bottleneck, as detailed in the following subsections.

\subsection{Scalable and Stable Gradient Estimation}
\label{subsec:sampling_optimzation}

\paragraph{Stratified Trajectory Sampling.}
The masked diffusion process entails a discrete transition sequence $\boldsymbol{x}_T \to \dots \to \boldsymbol{x}_0$. Naively sampling a single random timestep $t$ yields high-variance gradients because the denoising difficulty varies non-uniformly across the generation stages. To address this, we propose \textit{Stratified Trajectory Sampling} to guarantee dense temporal coverage (see Appendix~\ref{app:training_costs_ablations} for empirical validation of its gradient variance reduction). Specifically, for a trajectory of length $L$, we partition the valid timestep range into $K$ uniform segments. During each training step, we sample exactly one timestep $t_k$ from each segment:
\begin{equation}
\begin{split}
    t_k \sim \mathcal{U}\left[ \lfloor k \cdot \frac{L}{K} \rfloor, \lfloor (k+1) \cdot \frac{L}{K} \rfloor - 1 \right], \\ \quad k=0,\dots,K-1.
\end{split}
\label{eq:sampling}
\end{equation}

\paragraph{Block-wise Gradient Accumulation.}
Estimating an accurate velocity direction to rectify the policy requires aggregating statistics from multiple trajectories. For a batch size $B$, we sample $N$ trajectories per prompt and calculate gradients at $K$ stratified timesteps per trajectory. This produces an intractably large computational graph of size $B \times N \times K$. To resolve this memory bottleneck, we implement \textit{Block-wise Gradient Accumulation} (see Appendix~\ref{app:negative_controls} for an analysis of its critical role in memory feasibility). We partition the total $B \times N \times K$ samples into smaller, memory-efficient blocks. The optimization follows a hybrid parallel-serial execution: we compute gradients in parallel within each block to leverage GPU parallelism, then serially accumulate these updates across blocks. This technique allows us to substantially scale the effective batch size under fixed hardware constraints, thereby minimizing the variance of the policy gradient.

\begin{algorithm}[tb]
   \caption{Likelihood-Free Policy Optimization (\ours)}
   \label{alg:lfpo}
\begin{algorithmic}[1]
   \STATE {\bfseries Input:} Dataset $\mathcal{D}$, Initial Policy $\pi_{\theta}$, Params $\beta, N, K, \eta, \alpha$.
   \STATE {\bfseries Initialize:} Reference model parameters $\theta_{\text{old}} \leftarrow \theta$.
   \REPEAT
       \STATE \COMMENT{\textbf{Phase 1: Generate \& Estimate Rewards}}
       \STATE For batch queries $Q \subset \mathcal{D}$, sample $N$ trajectories $\mathcal{Y}_q \sim \pi_{\text{old}}(\cdot|q)$.
       \STATE Compute rewards $r_{\boldsymbol{y}} = \text{reward}(\boldsymbol{y})$ for all $\boldsymbol{y}$.
       \STATE $\mathcal{D}_{\text{batch}} \leftarrow \bigcup_{q \in Q} \{ (\boldsymbol{y}, r_{\boldsymbol{y}}) \mid \boldsymbol{y} \in \mathcal{Y}_q \}$.
       
       \STATE \COMMENT{\textbf{Phase 2: Block-wise Rectified Optimization}}
       \STATE Partition $\mathcal{D}_{\text{batch}}$ into blocks $\{\mathcal{B}_1, \dots, \mathcal{B}_M\}$.
       \FOR{each block $\mathcal{B}_m$}
           \STATE Sample stratified timesteps $\mathcal{T}_{\boldsymbol{y}} = \{t_1, \dots, t_K\}$ for all $\boldsymbol{y} \in \mathcal{B}_m$.
           \STATE Compute block loss (in parallel):
           \STATE \quad $\begin{aligned}
               \mathcal{L}_{\mathcal{B}} = &\sum_{(\boldsymbol{y}, r) \in \mathcal{B}_m} \sum_{t \in \mathcal{T}_{\boldsymbol{y}}} \Big[ r \cdot \text{CE}(\text{sg}(\pi^+), \pi_\theta) \\
               & + (1-r) \cdot \text{CE}(\text{sg}(\pi^-), \pi_\theta) \Big]
           \end{aligned}$
           \STATE Update policy parameters: $\theta \leftarrow \theta - \eta \nabla_\theta \mathcal{L}_{\mathcal{B}}$.
           \STATE \textbf{Delete} computation graph to free VRAM.
       \ENDFOR
       
       \STATE \COMMENT{\textbf{Phase 3: Reference Update}}
       \STATE Update reference model via EMA: $\theta_{\text{old}} \leftarrow \alpha \theta_{\text{old}} + (1-\alpha) \theta$.
   \UNTIL{convergence}
\end{algorithmic}
\end{algorithm}
\vspace{-3mm}

\subsection{\ours}
\label{subsec:algorithm}

The complete training procedure of \ours{} is summarized in Algorithm~\ref{alg:lfpo}, and the unified hyperparameter configurations applied across all evaluated tasks are detailed in Appendix~\ref{app:training_hyperparams}. During the initial data collection phase, we sample $N$ diverse trajectories per prompt using the reference policy $\pi_{\text{old}}$ (line 4) and compute task-specific scalar rewards $r$ (line 5). To optimize the policy under strict memory constraints, we apply a block-wise strategy (lines 8-15). Specifically, gradients are computed in parallel across $K$ stratified timesteps within each partitioned block (lines 11-12), immediately followed by a parameter update (line 13) and explicit deletion of the computation graph to prevent VRAM overflow (line 14). Finally, to maintain training stability, the reference model $\pi_{\text{old}}$ is updated via an \textit{Exponential Moving Average} (EMA) with decay rate $\alpha$ at the end of each iteration (line 17).

\section{Empirical Results}
\label{sec:results}

In this section, we provide a comprehensive evaluation of \ours{} across code generation and mathematical reasoning domains to validate its empirical performance and the efficiency gains inherent to its likelihood-free design. Section~\ref{subsec:setup} details the experimental setup. Section~\ref{subsec:main_results} analyzes the method's impact on performance, inference latency, and stability, followed by an in-depth ablation study of the contrastive velocity alignment in Section~\ref{subsec:ablation}.

\subsection{Experimental Setup}
\label{subsec:setup}

We evaluate \ours{} using DiffuCoder~\cite{gong2025diffucoderunderstandingimprovingmasked} for code and LLaDA 8B~\cite{nie2025large} for reasoning. We benchmark against state-of-the-art RL baselines, including Diffu/Coupled-GRPO~\cite{zhao2025d,gong2025diffucoderunderstandingimprovingmasked}, UniGRPO~\cite{yang2025mmada}, SPG~\cite{wang2025spgsandwichedpolicygradient}, and AGRPO~\cite{zhan2025principledtractablerlreasoning}. To ensure statistical rigor, all main results are reported as the mean and standard deviation across five independent random seeds.

\begin{table*}[t]
    \centering
    \caption{\textbf{Main Results on Code Generation Benchmarks.} We compare \textbf{\ours{}} against the base model DiffuCoder and various reinforcement learning baselines. Values in \textcolor{red}{red parentheses} denote absolute improvements over the base model. Standard deviations (\tiny{$\pm$}\small) across 5 random seeds are reported for the strongest baseline and our method to demonstrate statistical stability. \textbf{\ours{} (Pos. Only)} and \textbf{\ours{} (Neg. Only)} denote ablation variants optimized solely with the positive or negative loss, respectively, while \textbf{\ours{} (All Loss)} utilizes the full contrastive velocity rectification objective. Best results are highlighted in \textbf{bold}, and second-best are \underline{underlined}.}
    \label{tab:main_results}
    
    \newcommand{\inc}[1]{\small\textcolor{red}{(+#1)}}
    \newcommand{\dec}[1]{\small\textcolor{red}{(#1)}} 
    \newcommand{\std}[1]{\scriptsize{$\pm$#1}}
    
    \resizebox{\textwidth}{!}{%
    \begin{tabular}{lccccccc|c}
        \toprule
        \textbf{Model} & \textbf{HumanEval} & \textbf{HumanEval+} & \textbf{MBPP} & \textbf{MBPP+} & \textbf{EvalPlus} & \textbf{BigCodeBench} & \textbf{BigCodeBench} & \textbf{Avg.} \\
         & & & & & & \textbf{Full} & \textbf{Hard} & \\
        \midrule
        DiffuCoder & 72.0 & 65.2 & 75.1 & 61.9 & 63.6 & 35.7 & 12.2 & 55.1 \\
        \midrule
        + diffu-GRPO 
            & 73.1 \inc{1.1} & 66.1 \inc{0.9} & 77.3 \inc{2.2} & 67.3 \inc{5.4} & 67.9 \inc{4.3} & 39.2 \inc{3.5} & 12.5 \inc{0.3} & 57.6 \inc{2.5} \\
        + UniGRPO 
            & 74.5 \inc{2.5} & 67.2 \inc{2.0} & 78.9 \inc{3.8} & 68.8 \inc{6.9} & \textbf{70.3} \inc{6.7} & 42.6 \inc{6.9} & \underline{12.9} \inc{0.7} & 59.3 \inc{4.2} \\
        + SPG 
            & 74.6 \inc{2.6} & 69.1 \inc{3.9} & 80.2 \inc{5.1} & 69.2 \inc{7.3} & 68.9 \inc{5.3} & 42.1 \inc{6.4} & \underline{12.9} \inc{0.7} & 59.6 \inc{4.5} \\
        + coupled-GRPO 
            & 73.2 \inc{1.2} & 68.3 \inc{3.1} & 78.6 \inc{3.5} & 67.5 \inc{5.6} & 67.9 \inc{4.3} & 40.4 \inc{4.7} & 10.8 \dec{-1.4} & 58.1 \inc{3.0} \\
        + AGRPO 
            & \underline{75.3}\std{1.1} \inc{3.3} & \textbf{70.2}\std{0.9} \inc{5.0} & \underline{80.3}\std{1.0} \inc{5.2} & \textbf{71.7}\std{1.2} \inc{9.8} & 69.3\std{0.8} \inc{5.7} & \underline{44.6}\std{1.1} \inc{8.9} & \textbf{13.1}\std{0.9} \inc{0.9} & \underline{60.6}\std{1.0} \inc{5.5} \\
        \midrule
        \textbf{+ \ours{} (Pos. Only)}
            & 73.2 \inc{1.2} & 67.3 \inc{2.1} & 80.1 \inc{5.0} & 68.2 \inc{6.3} & 68.3 \inc{4.7} & 43.1 \inc{7.4} & 12.4 \inc{0.2} & 58.9 \inc{3.8} \\
        \textbf{+ \ours{} (Neg. Only)}
            & 73.1 \inc{1.1} & 67.2 \inc{2.0} & 79.5 \inc{4.4} & 68.5 \inc{6.6} & 68.9 \inc{5.3} & 43.4 \inc{7.7} & 12.5 \inc{0.3} & 59.0 \inc{3.9} \\
        \textbf{+ \ours{} (All Loss)} 
            & \textbf{75.6}\std{0.2} \inc{3.6} & \underline{70.1}\std{0.3} \inc{4.9} & \textbf{81.6}\std{0.2} \inc{6.5} & \underline{71.3}\std{0.3} \inc{9.4} & \underline{69.5}\std{0.2} \inc{5.9} & \textbf{44.8}\std{0.3} \inc{9.1} & \underline{12.9}\std{0.2} \inc{0.7} & \textbf{60.8}\std{0.2} \inc{5.7} \\
        \bottomrule
    \end{tabular}
    }
\end{table*}

\begin{table*}[t]
    \centering
    \caption{\textbf{Main Results on Reasoning Benchmarks.} We compare \textbf{\ours{}} against the base model LLaDA 8B and various reinforcement learning baselines. Formatting conventions, standard deviation reporting across 5 random seeds, and the nomenclature for ablation variants strictly follow those detailed in Table~\ref{tab:main_results}.}
    \label{tab:reasoning_results}
    
    \newcommand{\inc}[1]{\small\textcolor{red}{(+#1)}}
    \newcommand{\dec}[1]{\small\textcolor{red}{(#1)}} 
    \newcommand{\std}[1]{\scriptsize{$\pm$#1}}
    
    \resizebox{\textwidth}{!}{%
    \begin{tabular}{lcccccc|c}
        \toprule
        \textbf{Model} & \textbf{GSM8K} & \textbf{MATH} & \textbf{GPQA} & \textbf{Hellaswag} & \textbf{WinoGrande} & \textbf{PIQA} & \textbf{Avg.} \\
        \midrule
        LLaDA 8B & 69.7 & 30.6 & 25.3 & 70.3 & 75.8 & 71.6 & 57.2 \\
        \midrule
        + diffu-GRPO 
            & 74.3 \inc{4.6} & 33.7 \inc{3.1} & 26.1 \inc{0.8} & 80.1 \inc{9.8} & 83.2 \inc{7.4} & 79.3 \inc{7.7} & 62.8 \inc{5.6} \\
        + UniGRPO 
            & 77.2 \inc{7.5} & 34.5 \inc{3.9} & 26.2 \inc{0.9} & 82.3 \inc{12.0} & 84.7 \inc{8.9} & 81.2 \inc{9.6} & 64.4 \inc{7.2} \\
        + SPG 
            & 76.2 \inc{6.5} & 34.1 \inc{3.5} & 25.9 \inc{0.6} & 83.1 \inc{12.8} & 84.3 \inc{8.5} & 80.9 \inc{9.3} & 64.1 \inc{6.9} \\
        + coupled-GRPO 
            & 75.1 \inc{5.4} & 33.9 \inc{3.3} & 25.7 \inc{0.4} & 82.5 \inc{12.2} & 85.6 \inc{9.8} & 83.2 \inc{11.6} & 64.3 \inc{7.1} \\
        + AGRPO 
            & \underline{78.3}\std{1.3} \inc{8.6} & \underline{36.1}\std{0.9} \inc{5.5} & \textbf{27.3}\std{1.1} \inc{2.0} & \underline{85.1}\std{1.0} \inc{14.8} & \textbf{87.3}\std{1.2} \inc{11.5} & \underline{85.6}\std{0.8} \inc{14.0} & \underline{66.6}\std{1.0} \inc{9.4} \\
        \midrule
        \textbf{+ \ours{} (Pos. Only)} 
            & \underline{78.3} \inc{8.6} & 35.2 \inc{4.6} & 26.1 \inc{0.8} & 84.2 \inc{13.9} & 83.1 \inc{7.3} & 80.6 \inc{9.0} & 64.6 \inc{7.4} \\
        \textbf{+ \ours{} (Neg. Only)} 
            & 78.1 \inc{8.4} & 34.9 \inc{4.3} & 26.3 \inc{1.0} & 84.3 \inc{14.0} & 83.7 \inc{7.9} & 79.6 \inc{8.0} & 64.5 \inc{7.3} \\
        \textbf{+ \ours{} (All Loss)} 
            & \textbf{79.6}\std{0.3} \inc{9.9} & \textbf{37.6}\std{0.2} \inc{7.0} & \underline{27.1}\std{0.3} \inc{1.8} & \textbf{85.7}\std{0.2} \inc{15.4} & \underline{86.9}\std{0.2} \inc{11.1} & \textbf{85.9}\std{0.2} \inc{14.3} & \textbf{67.1}\std{0.2} \inc{9.9} \\
        \bottomrule
    \end{tabular}
    }
    \vspace{-2mm}
\end{table*}

\paragraph{Benchmarks and Data Strategy.}
We employ distinct evaluation protocols for each domain. For code generation, we target \textit{Out-of-Domain Generalization}: models are trained on AceCode-87K~\cite{zeng-etal-2025-acecoder} and evaluated zero-shot on HumanEval~\cite{chen2021evaluatinglargelanguagemodels}, MBPP~\cite{austin2021programsynthesislargelanguage}, EvalPlus~\cite{evalplus}, and BigCodeBench~\cite{zhuo2025bigcodebench}. For mathematical reasoning, we follow standard \textit{In-Domain Evaluation} using the training and test splits of GSM8K~\cite{cobbe2021trainingverifierssolvemath} and MATH~\cite{hendrycksmath2021}, alongside general evaluations on GPQA~\cite{rein2024gpqa} and PIQA~\cite{bisk2019piqareasoningphysicalcommonsense}.

\paragraph{Implementation and Inference Protocol.}
Our framework utilizes sparse binary rewards, specifically functional correctness for code and symbolic accuracy for reasoning. We optimize the policy using AdamW combined with our memory-efficient block-wise strategy. For the LLaDA reasoning experiments, we directly RL-tune the LLaDA 8B base checkpoint to test whether the proposed objective can improve a non-instruction-tuned dLLM. We use temperature $0$, maximum generated answer length $512$, a CoT prompt template ``Let's think step by step'' followed by the question, and the standard LLaDA answer extraction/evaluation protocol. This checkpoint choice differs from high-performing instruction-tuned or separately post-trained LLaDA variants reported in concurrent work, and is therefore not intended as a direct checkpoint-to-checkpoint comparison. To guarantee fair comparisons among the diffusion baselines in our experiments, we strictly standardize the inference protocol. This unified protocol features a Cosine noise schedule with a maximum of $T=2048$ steps, a schedule-driven parallel update rule to maximize hardware utilization, and a confidence-based early stopping mechanism with a threshold of $\tau=0.99$ and a patience of $3$ steps. Comprehensive training hyperparameter configurations and the detailed inference protocol are provided in Appendix~\ref{app:training_hyperparams} and Appendix~\ref{app:inference_hyperparams}, respectively.

\subsection{Main Results}
\label{subsec:main_results}

\paragraph{Performance and Stability.}
We evaluate generation quality across downstream tasks to directly validate the core advantage of our likelihood-free design (Tables~\ref{tab:main_results} and~\ref{tab:reasoning_results}). \ours{} consistently outperforms the strongest likelihood-based baseline, AGRPO, achieving the highest average scores in both code generation (60.8) and reasoning (67.1). Notably, it yields substantial absolute improvements over the base models on foundational benchmarks, including HumanEval (+3.6\%), GSM8K (+9.9\%), and MATH (+7.0\%). We attribute this superior accuracy to our token-level velocity alignment, which provides a direct learning signal and avoids the high-variance importance sampling inherent to traditional RL methods. Crucially, bypassing sequence likelihood estimation yields unprecedented optimization stability. Across five independent random seeds, \ours{} maintains an exceptionally low standard deviation ($\pm 0.2$ to $\pm 0.3$), starkly contrasting with the high volatility of AGRPO ($\pm 0.8$ to $\pm 1.3$). This confirms that our localized empirical surrogate provides a fundamentally more reliable training paradigm (see Appendix~\ref{app:statistical_reporting} for formal statistical analysis).

\begin{table*}[t]
    \centering
    \caption{\textbf{Unified Efficiency Analysis.} We report the average inference steps per problem across code generation and reasoning benchmarks. \textbf{Avg.} represents the mean steps within each task category. The values in \textcolor{red}{red parentheses} denote the reduction in inference steps (efficiency improvement) compared to the Base Model. \textit{Base Model} refers to DiffuCoder for code tasks and LLaDA 8B for reasoning tasks. Lower values ($\downarrow$) indicate better efficiency.}
    \label{tab:unified_efficiency}
    
    \newcommand{\dec}[1]{\small\textcolor{red}{(-#1)}}
    \newcommand{\inc}[1]{\small{(+#1)}} 
    
    \resizebox{\textwidth}{!}{%
    \begin{tabular}{lcccc|cccc}
        \toprule
        \multirow{2}{*}{\textbf{Model}} & \multicolumn{4}{c|}{\textbf{Code Generation}} & \multicolumn{4}{c}{\textbf{Reasoning}} \\
        \cmidrule(lr){2-5} \cmidrule(lr){6-9}
         & \textbf{HumanEval} & \textbf{MBPP} & \textbf{BCB-Full} & \textbf{Avg.} & \textbf{GSM8K} & \textbf{MATH} & \textbf{Hellaswag} & \textbf{Avg.} \\
        \midrule
        Base Model$^\dagger$ & 253.7 & 317.2 & 426.3 & 332.4 & 1034.2 & 1536.3 & 1036.7 & 1202.4 \\
        \midrule
        + diffu-GRPO 
            & 254.3 \inc{0.6} & 305.4 \dec{11.8} & 430.7 \inc{4.4} & 330.1 \dec{2.3} & 1048.4 \inc{14.2} & 1547.4 \inc{11.1} & 1027.8 \dec{8.9} & 1207.9 \inc{5.5} \\
        + UniGRPO 
            & 248.0 \dec{5.7} & 319.3 \inc{2.1} & 443.3 \inc{17.0} & 336.9 \inc{4.5} & 1020.3 \dec{13.9} & 1539.3 \inc{3.0} & 985.8 \dec{50.9} & 1181.8 \dec{20.6} \\
        + SPG 
            & 243.7 \dec{10.0} & 305.8 \dec{11.4} & 416.7 \dec{9.6} & 322.1 \dec{10.3} & 1071.0 \inc{36.8} & 1576.7 \inc{40.4} & 1009.1 \dec{27.6} & 1218.9 \inc{16.5} \\
        + coupled-GRPO 
            & 262.2 \inc{8.5} & 326.5 \inc{9.3} & 431.2 \inc{4.9} & 340.0 \inc{7.6} & 1015.9 \dec{18.3} & 1601.6 \inc{65.3} & 1025.8 \dec{10.9} & 1214.4 \inc{12.0} \\
        + AGRPO 
            & 254.2 \inc{0.5} & 313.9 \dec{3.3} & 438.5 \inc{12.2} & 335.5 \inc{3.1} & 1004.8 \dec{29.4} & 1609.9 \inc{73.6} & 1063.6 \inc{26.9} & 1226.1 \inc{23.7} \\
        \midrule
        \textbf{+ \ours{} (Pos. Only)} 
            & \underline{223.6} \dec{30.1} & 292.3 \dec{24.9} & \underline{373.4} \dec{52.9} & \underline{296.4} \dec{36.0} & 911.3 \dec{122.9} & \underline{1355.3} \dec{181.0} & 893.1 \dec{143.6} & 1053.2 \dec{149.2} \\
        \textbf{+ \ours{} (Neg. Only)} 
            & 234.7 \dec{19.0} & \underline{286.7} \dec{30.5} & 377.3 \dec{49.0} & 299.6 \dec{32.8} & \underline{909.7} \dec{124.5} & 1367.9 \dec{168.4} & \underline{891.2} \dec{145.5} & \underline{1056.3} \dec{146.1} \\
        \textbf{+ \ours{} (All Loss)} 
            & \textbf{218.3} \dec{35.4} & \textbf{282.3} \dec{34.9} & \textbf{371.2} \dec{55.1} & \textbf{290.6} \dec{41.8} & \textbf{891.2} \dec{143.0} & \textbf{1348.9} \dec{187.4} & \textbf{890.2} \dec{146.5} & \textbf{1043.4} \dec{159.0} \\
        \bottomrule
    \end{tabular}
    }
\end{table*}

\begin{figure*}[t]

\centering
\includegraphics[width=1.0\textwidth]{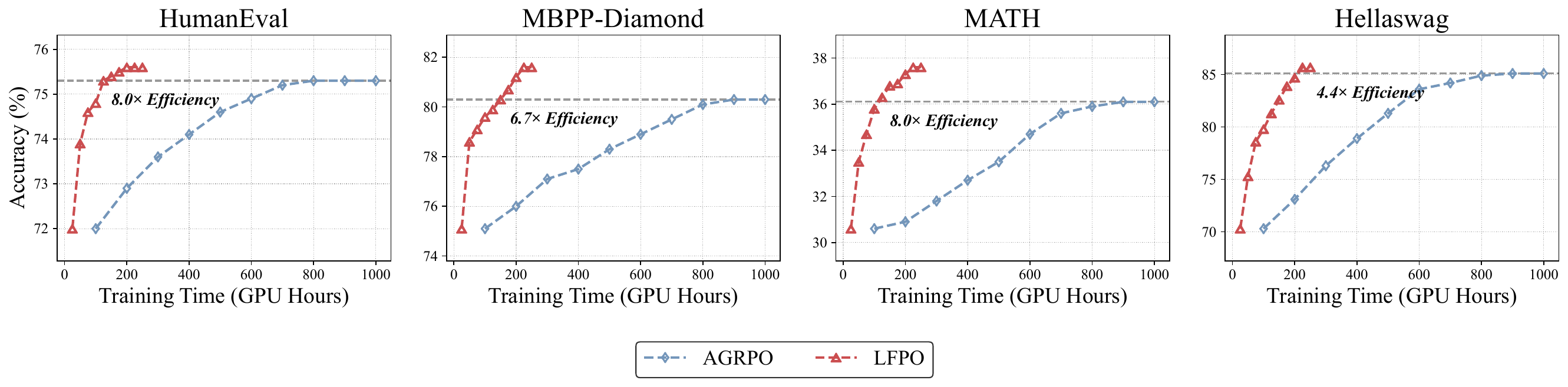}

\caption{\textbf{Convergence Analysis on Code and Reasoning Tasks.} We plot accuracy progression against training time under the matched hardware setting. The red and blue curves denote our proposed \ours{} and the AGRPO baseline, respectively. The horizontal dashed line marks the baseline's final converged accuracy. Notably, \ours{} requires less time to match or surpass this peak performance, underscoring \ours{} sample efficiency and rapid convergence.}
\label{fig:convergence}
\vspace{-2mm}
\end{figure*}

\paragraph{Inference Efficiency via Straighter Trajectory Learning.}
A critical bottleneck for diffusion language models is the high computational overhead of iterative denoising. Table~\ref{tab:unified_efficiency} highlights a definitive advantage of our approach: \ours{} significantly accelerates inference while concurrently elevating accuracy. Compared to the base models, our method reduces the average inference steps by 41.8 steps for code generation and 159.0 steps for reasoning tasks. In stark contrast, likelihood-based baselines like AGRPO often degrade efficiency (e.g., increasing steps by 73.6 on MATH) to achieve marginal performance gains. This divergence stems from a fundamental difference in optimization objectives. Standard likelihood maximization implicitly forces models to mimic specific, often meandering denoising trajectories observed in the training data. Conversely, by casting generation as a geometric velocity alignment problem, \ours{} encourages the model to learn a straightened probability flow from the mask prior to the target data distribution. This empirical straightening actively shortens the generation trajectory, enabling the model to converge to high-quality solutions with drastically fewer intermediate steps (see Appendix~\ref{sec:case_study} for a visual case study).

\paragraph{Training Convergence and Deconstructing Acceleration.}
Beyond inference efficiency, Figure~\ref{fig:convergence} highlights the rapid training convergence of our framework. Compared to the AGRPO baseline, \ours{} matches peak performance significantly faster under the same hardware setting ($8.0\times$ on HumanEval and MATH, $6.7\times$ on MBPP-Diamond, and $4.4\times$ on HellaSwag) (detailed hardware configurations, wall-clock costs, throughput metrics, and a JustGRPO training comparison are provided in Appendix~\ref{app:hardware}). We attribute this acceleration to two synergistic factors. First, our contrastive dual-policy design yields high sample efficiency. By anchoring intermediate states to high-reward terminal predictions, this push-and-pull mechanism encourages straightened probability flows, which enables the model to reach correct solutions in fewer iterative steps and accelerates data collection. Second, our tailored gradient estimation techniques (Section~\ref{subsec:sampling_optimzation}) optimize both update stability and computational throughput. Specifically, Stratified Trajectory Sampling guarantees dense temporal coverage to reduce gradient variance, providing a highly stable optimization signal per step. Concurrently, block-wise parallelization minimizes computational overhead by operating directly in the discrete logit space to update multiple tokens simultaneously, avoiding the poorly scaling sequential backtracking of traditional methods. Ultimately, the combination of high sample efficiency, low-variance updates, and parallelized computation drives the superior wall-clock convergence observed across our evaluations.

\subsection{Ablation Study and Sensitivity Analysis}
\label{subsec:ablation}

\paragraph{Ablation on Contrastive Components.}
To dissect the geometric mechanisms driving \ours{}, we analyze variants optimized with partial objectives: \textit{Pos. Only} (Attraction) and \textit{Neg. Only} (Repulsion). As shown in Tables~\ref{tab:main_results} and~\ref{tab:reasoning_results}, both variants yield improvements over the base models, yet neither matches the overall performance of the complete \textit{All Loss} objective. From a geometric perspective on the probability simplex (Figure~\ref{fig:method}), the \textit{Pos. Only} term acts as an attraction force, pulling the model's velocity $v_\theta^{(o)}$ toward the vertex of the correct token $\boldsymbol{x}_1$. Conversely, the \textit{Neg. Only} term acts as a repulsion force, pushing the velocity away from incorrect vertices. The superior performance of the combined objective suggests that shaping the vector field benefits from a contrastive approach: simultaneously encouraging correct directions while suppressing deviations into low-reward regions yields the most reliable generative flow.

\paragraph{Sensitivity Analysis and Case Studies.}
Extensive sensitivity analyses (Appendix~\ref{app:sensitivity}) dissect the empirical gains of \ours{} across various configurations. First, ablations on core hyperparameters and negative controls (Appendices~\ref{app:training_costs_ablations} and \ref{app:negative_controls}) demonstrate that our block-wise accumulation strictly bounds peak VRAM, while validating the absolute necessity of logit-space perturbation for optimization stability. Second, evaluations of inference decoding parameters (Appendix~\ref{app:inference_decoding}) reveal that a Cosine masking schedule combined with a high confidence threshold ($\tau=0.99$) yields an optimal accuracy-efficiency balance, slashing denoising steps without degrading quality. Third, generalization tests on the Dream architecture (Appendix~\ref{app:generalization}) confirm that our contrastive alignment framework is architecture-agnostic and universally applicable. Finally, qualitative case studies (Appendix~\ref{sec:case_study}) visually corroborate our core theoretical claim, illustrating exactly how \ours{} anchors to correct terminal states early to straighten probability flows and bypass meandering denoising steps.

\section{Conclusion}
\label{sec:conclusion}

We introduced \ours{}, a likelihood-free alignment framework for discrete diffusion language models that bypasses high-variance RL likelihood estimation via contrastive velocity alignment. Evaluations on reasoning and coding tasks demonstrate that \ours{} concurrently delivers superior accuracy, optimization stability, and inference acceleration.

\newpage

\section*{Limitations}
\label{sec:limitations}

While \ours{} presents a highly effective empirical framework, we acknowledge several limitations that warrant future investigation.

First, the method is constrained by a cold-start bottleneck. Because our contrastive optimization relies on anchoring to high-reward terminal states, the base model must possess a non-zero zero-shot success rate to discover initial valid trajectories; \ours{} propagates sequence-level verifier rewards to token states visited by sampled trajectories, but does not provide exact per-token credit assignment.
Second, although block-wise parallelization mitigates VRAM constraints, the overall training process maintains a substantial computational cost due to the necessity of generating $N$ rollouts and evaluating $K$ stratified timesteps per trajectory. 
Third, the current framework is exclusively tailored for verifiable sparse reward scenarios (e.g., exact mathematical matching and code execution). It does not trivially extend to subjective alignment tasks governed by dense human preference models. 
Fourth, unlike classical reinforcement learning algorithms, \ours{} operates as an empirical velocity surrogate and does not currently provide strict theoretical guarantees for monotonic policy improvement or global convergence. 
Finally, despite our reference-anchored design mitigating certain reward over-optimization risks, the potential for reward hacking and benchmark overfitting remains a persistent challenge necessitating careful monitoring in practical deployments.

\newpage
\medskip
\bibliography{custom}
\newpage

\appendix

\newpage
\appendix
\onecolumn
\appendix

\appendix

\section{Related Work} \label{sec:related_work}

\paragraph{Diffusion Large Language Models.}
Discrete denoising diffusion probabilistic models (D3PMs)~\cite{austin2021structured} have recently emerged as a compelling non-autoregressive paradigm for text generation~\cite{yu2025discretediffusionlargelanguage}. Unlike traditional autoregressive models that generate tokens strictly left-to-right, dLLMs generate text via parallel iterative unmasking, offering flexible bidirectional context modeling and potential improvements in decoding efficiency~\cite{ye2025dream}. Recent large-scale implementations, such as LLaDA~\cite{nie2025large} and DiffuCoder~\cite{gong2025diffucoderunderstandingimprovingmasked}, have demonstrated that dLLMs can achieve language modeling performance competitive with their autoregressive counterparts~\cite{fan2026stablediffcoderpushingfrontiercode}. However, while pre-training endows these models with strong general capabilities, effective alignment techniques to enhance their performance remain underexplored.

\paragraph{Reinforcement Learning for Diffusion Alignment.}
To bridge this gap, recent research has focused on adapting PG methods to the discrete diffusion setting. A prominent line of work utilizes the \textit{Group Relative Policy Optimization} (GRPO)~\cite{shao2024deepseekmathpushinglimitsmathematical} framework to enable critic-free optimization. 
Diffu-GRPO~\cite{zhao2025d} pioneered this direction, introducing the first PG algorithm tailored for masked dLLMs to improve mathematical reasoning. 
UniGRPO~\cite{yang2025mmada} extended this framework to multimodal domains, unifying reasoning and generation tasks with diversified reward modeling. 
To address the gradient bias caused by the intractable likelihood of diffusion trajectories, \textit{Sandwiched Policy Gradient} (SPG)~\cite{wang2025spgsandwichedpolicygradient} proposed leveraging both upper and lower likelihood bounds. 
More recently, Coupled-GRPO~\cite{gong2025diffucoderunderstandingimprovingmasked} and AGRPO~\cite{zhan2025principledtractablerlreasoning} have been proposed to further improve stability and sample efficiency. Specifically, AGRPO achieves strong results by employing Monte Carlo sampling for unbiased policy gradient estimation.
Concurrent sequence-level approaches further explore different trade-offs. JustGRPO~\cite{ni2026flexibilitytrap} applies vanilla GRPO with an autoregressive-style training strategy and reports strong results when substantially more compute and a different checkpoint are used. ESPO~\cite{ou2025principledsequence} and GDPO~\cite{rojas2025improvingreasoninggdpo} revisit dLLM RL from sequence-level perspectives, proposing alternative objectives for reasoning-oriented diffusion alignment. These works are complementary to our focus: \ours{} targets the likelihood bottleneck by replacing sequence-level likelihood estimation with a detached token-level logit surrogate.
The distinction between AGRPO and \ours{} is therefore not that AGRPO lacks Monte Carlo estimation, but that the two methods estimate different objects. AGRPO remains within a trajectory-level policy-gradient framework and uses Monte Carlo samples to estimate policy-gradient quantities for diffusion trajectories. \ours{} instead constructs detached token-level implicit targets and optimizes a local CE/logit surrogate motivated by the shared-origin velocity residual. This changes the optimization signal from trajectory likelihood estimation to token-level vector-field rectification, which is where our observed variance and efficiency gains arise.

Despite these advancements, the aforementioned methods predominantly remain within the paradigm of maximizing the likelihood (or its surrogates) of high-reward trajectories. We argue that strictly adhering to this likelihood-based objective creates a bottleneck for dLLMs. Rather than persisting with likelihood maximization, we posit that dLLMs require a native RL formulation grounded in their geometric nature. In the continuous domain, FM~\cite{lipman2023flow, liu2023flow} has successfully replaced likelihood objectives with stable vector field regression. While translating this to the discrete domain is non-trivial, recent theoretical works~\cite{zheng2026diffusionnft, liu2025flowgrpo} have begun to uncover the geometric structures of discrete diffusion. 
Building on these insights, our work identifies a pointwise logit-gradient connection between token-level CE updates and discrete velocity residuals. This perspective motivates us to reframe alignment not as trajectory probability maximization, but as velocity rectification, leading to the stable and memory-efficient optimization strategy that we introduce in the following section.

\section{Proof of Theorem \ref{thm:gradient_alignment}}
\label{app:gradient_derivation}

\begin{proof}
We analyze the gradient of the Cross-Entropy loss $\mathcal{L}_{CE}$ with respect to the $i$-th component of the pre-softmax logits, $z_i$. Let $p_k = \frac{\exp(z_k)}{\sum_{j=1}^V \exp(z_j)}$ and the loss be $\mathcal{L}_{CE} = -\sum_{j=1}^V x_{1,j} \log p_j$. Applying the chain rule, we expand the partial derivative:
\begin{align}
    \frac{\partial \mathcal{L}_{CE}}{\partial z_i} &= \sum_{j=1}^V \frac{\partial \mathcal{L}_{CE}}{\partial p_j} \frac{\partial p_j}{\partial z_i} \\
    &= \sum_{j=1}^V \left[ - \frac{x_{1,j}}{p_j} \right] \left[ p_j (\delta_{ij} - p_i) \right] \\
    &= - \sum_{j=1}^V x_{1,j} (\delta_{ij} - p_i) \\
    &= - \left[ \sum_{j=1}^V x_{1,j} \delta_{ij} - \sum_{j=1}^V x_{1,j} p_i \right] \\
    &= - \left[ x_{1,i} - p_i \sum_{j=1}^V x_{1,j} \right]
\end{align}
Since $\boldsymbol{x}_1$ is a valid one-hot probability distribution ($\sum_{j=1}^V x_{1,j} = 1$), the scalar derivative simplifies to $\frac{\partial \mathcal{L}_{CE}}{\partial z_i} = p_i - x_{1,i}$. Expressing this in vector notation over the vocabulary $V$, let $\boldsymbol{o}_t \in \Delta^{V-1}$ denote any shared origin fixed for the current input state (e.g., the mask prior $\boldsymbol{m}$ or the current denoising state $\boldsymbol{x}_t$). Define the model velocity field $v_\theta^{(o)} = \boldsymbol{p} - \boldsymbol{o}_t$ and the target velocity field $u_t^{(o)} = \boldsymbol{x}_1 - \boldsymbol{o}_t$. Their residual cancels the shared origin:
\begin{align}
    v_\theta^{(o)} - u_t^{(o)} &= (\boldsymbol{p} - \boldsymbol{o}_t) - (\boldsymbol{x}_1 - \boldsymbol{o}_t) \\
    &= \boldsymbol{p} - \boldsymbol{x}_1 \\
    &= \nabla_{\boldsymbol{z}} \mathcal{L}_{CE}
\end{align}
This yields the pointwise logit-gradient identity. \end{proof}

\section{Additional Method Details}
\label{app:method_details}

In this section, we provide supplementary discussions to deeply contextualize the algorithmic design choices of \ours{} within the broader landscape of diffusion reinforcement learning and discrete policy optimization. 

\paragraph{Methodological Positioning and Contributions.}
Table~\ref{tab:contribution_decomposition} delineates the conceptual boundaries between \ours{} and closely related paradigms. While continuous diffusion RL and flow-matching frameworks (e.g., DiffusionNFT, Flow-GRPO) inspire our positive and negative target formulation, they inherently operate in continuous spaces and do not address discrete sequence likelihoods. Conversely, standard dLLM reinforcement learning methods like AGRPO rely on policy-gradient estimators that require approximating high-variance trajectory likelihoods. \ours{} bridges this gap by adapting the contrastive target into a discrete logit-space objective. This unique instantiation provides a token-level surrogate loss that completely circumvents trajectory likelihood estimation. Furthermore, we support this theoretical adaptation with a highly scalable block-wise parallelization strategy and a standardized evaluation protocol for reproducible benchmarking.

\begin{table}[h]
\centering
\small
\caption{\textbf{Positioning of \ours{} relative to closely related RL/FM methods.} The positive--negative implicit target idea is inspired by prior diffusion RL/FM work. The main contribution of \ours{} is the discrete dLLM instantiation: token-level CE/logit supervision that avoids sequence-likelihood estimation, together with scalable sampling/optimization and standardized dLLM RLVR evaluation.}
\label{tab:contribution_decomposition}
\begin{tabularx}{\textwidth}{p{0.24\textwidth}XXX}
\toprule
\textbf{Component} & \textbf{DiffusionNFT / Flow-GRPO} & \textbf{AGRPO} & \textbf{\ours{}} \\
\midrule
Positive--negative or reward-induced target direction & Inspired by this line of diffusion RL/FM work~\cite{zheng2026diffusionnft,liu2025flowgrpo} & Not the central design & Adopted as a detached contrastive target in discrete logit space \\
Discrete masked-dLLM token-level formulation & Continuous/image or flow-matching setting & Likelihood-based dLLM policy-gradient estimator & Token-level CE/logit surrogate for masked dLLMs \\
Sequence/trajectory likelihood handling & Not designed for masked dLLM sequence likelihood & Estimates or approximates policy-gradient quantities & Avoids sequence/trajectory likelihood estimation; retains token-level CE/logit supervision \\
Perturbation space & Model/flow-space target construction & Policy-gradient style update & Logit-space perturbation to keep simplex constraints smooth \\
Temporal credit assignment & Method-specific continuous-time sampling & Trajectory-level PG sampling & Stratified timestep sampling over denoising trajectories \\
Scalable implementation & Not focused on dLLM token-logit memory & Monte Carlo estimator for dLLMs & Block-wise logit parallelization for $B \times N \times K$ training samples \\
Evaluation scope & Diffusion RL/FM benchmarks & dLLM RL baselines & Standardized dLLM RLVR evaluation on code and reasoning with shared decoding protocol \\
\bottomrule
\end{tabularx}
\end{table}

\paragraph{Geometric Interpretation vs. Naive Cross-Entropy.}
Table~\ref{tab:fm_vs_ce} details why the geometric flow-matching perspective is fundamentally superior to a naive reward-weighted cross-entropy (CE) approach. Standard reward-weighted CE essentially acts as an attraction-only heuristic, pulling the model exclusively toward sampled high-reward trajectories. By framing the generation process as a shared-origin velocity alignment problem, \ours{} formally constructs explicit positive and negative implicit policies ($\pi^+$ and $\pi^-$). This contrastive formulation actively pulls the generation toward high-reward directions while simultaneously repelling it from low-reward deviations. As demonstrated by the performance gap between the \textit{Pos. Only} variant and the complete \textit{All Loss} variant in our ablation studies, this dual push-and-pull mechanism is vital for maintaining a reliable generative flow.

\begin{table}[h]
\caption{\textbf{Why the FM view changes the implementation.} The algebraic CE gradient gives a tractable token-level update, while the shared-origin velocity view motivates the contrastive target construction used by \ours{}.}
\label{tab:fm_vs_ce}
\centering
\small
\begin{tabularx}{0.86\textwidth}{lXX}
\toprule
\textbf{Aspect} & \textbf{Naive reward-weighted CE} & \textbf{\ours{}} \\ \midrule
Target construction & Pull toward sampled high-reward outputs & Detached positive/negative implicit policies $(\pi^+,\pi^-)$ \\
Geometric view & Supervised attraction heuristic & Shared-origin residual rectification \\
Update effect & Attraction-only signal & Pull high-reward directions and push away low-reward directions \\
Empirical check & Pos. Only variant & All Loss variant in Tables~\ref{tab:main_results}--\ref{tab:reasoning_results} \\
\bottomrule
\end{tabularx}
\end{table}

\paragraph{Simplex Constraints and Logit-Space Perturbation.}
A critical technical detail of \ours{} is the space in which the policy perturbation is applied, as compared in Table~\ref{tab:perturbation_comparison}. Applying linear perturbations directly in the probability space violates fundamental simplex constraints (i.e., $p_i \in [0,1]$ and $\sum p_i = 1$). Enforcing these constraints post hoc requires harsh clipping or truncation operations, which severely fragment gradient continuity and degrade optimization stability. By constructing targets and applying perturbations strictly in the unconstrained logit space ($z_i \in \mathbb{R}$), \ours{} preserves smooth relative updates. The valid probability simplex is then naturally recovered via the Softmax operation without discarding crucial gradient information. This structural choice directly translates to empirical gains, elevating the HumanEval score from 73.9 (using probability clipping) to 75.6 (using our full logit-space formulation).

\begin{table}[h]
\caption{\textbf{Properties of different spaces for policy perturbation.} Probability-space perturbation uses clipping/truncation after a linear update, while \ours{} constructs targets in logit space and maps them back to the simplex.}
\label{tab:perturbation_comparison}
\centering
\small
\begin{tabularx}{0.92\textwidth}{lXXX}
\toprule
\textbf{Property} & \textbf{Probability Space} & \textbf{Log-Prob. Space} & \textbf{Logit Space (Ours)} \\ \midrule
Domain & $p_i\in[0,1]$ & $\log p_i\le 0$ & $z_i\in\mathbb{R}$ \\
Linear perturbation & Requires clipping & Shift-normalized & Unconstrained before Softmax \\
Simplex constraint & Broken before truncation & Recovered by normalization & Recovered by Softmax \\
Gradient continuity & Fragmented by clipping & Mostly smooth & Smooth relative update \\
HumanEval check & 73.9 with clipping & -- & 75.6 with full \ours{} \\
\bottomrule
\end{tabularx}
\end{table}

\section{Scope of the Theoretical Claim}
\label{app:theory_scope}

Theorem~\ref{thm:gradient_alignment} is intentionally pointwise. It states an identity in unconstrained logit space for a fixed input state and a fixed shared origin; it does not establish global equivalence between the CE objective and a continuous FM objective, nor does it prove monotonic policy improvement. The identity is useful because it gives a tractable local residual direction that avoids sequence/trajectory likelihood estimation.

One limited implication is that the token-level CE logit gradient is bounded on the probability simplex. Since $p_\theta$ and $\boldsymbol{x}_1$ are probability vectors and $\boldsymbol{x}_1$ is one-hot,
\begin{equation}
    \|\nabla_{\boldsymbol{z}}\mathcal{L}_{CE}\|_2
    = \|p_\theta-\boldsymbol{x}_1\|_2
    \le \sqrt{2}.
\end{equation}
This bounded token-level signal helps explain why \ours{} has lower observed cross-seed variance than trajectory-likelihood estimators in our experiments. However, this observation should be interpreted as motivation for an empirical low-variance surrogate, not as a formal convergence guarantee for sparse-reward RLVR.

\section{Experimental Details and Hyperparameters}
\label{app:hyperparams}

In this section, we provide the comprehensive hyperparameter configurations and implementation details for \ours{} to ensure reproducibility. All experiments were conducted using the PyTorch framework on a cluster equipped with NVIDIA H20 (96GB) GPUs.

\subsection{Training Configurations and Unified Hyperparameters}
\label{app:training_hyperparams}

To ensure full reproducibility and demonstrate the robust out-of-the-box generalization of \ours{}, we utilize a unified hyperparameter configuration across completely different tasks and model architectures (DiffuCoder for code generation and LLaDA 8B for mathematical reasoning). This design choice highlights that our framework does not rely on heavy, task-specific tuning to achieve strong performance.

We employ the AdamW optimizer with a linear learning rate warm-up for the first 10\% of training steps, followed by a cosine decay schedule. To manage the memory constraints of dLLMs during stratified trajectory sampling, we implement the Block-wise Gradient Accumulation as described in Section~\ref{sec:method}. 

Crucially, contrary to prior likelihood-based methods that often require brittle, domain-specific adjustments, \ours{} maintains consistent core parameters. The contrastive coefficient $\beta$ is fixed at 1.0, and we sample $N=10$ trajectories per prompt with $K=8$ stratified timesteps. The dynamic EMA decay rate $\alpha$ is scaled according to the training step to prevent early policy collapse. The complete, unified hyperparameter settings applied to all benchmarks are summarized in Table~\ref{tab:unified_hyperparams}.

\begin{table}[h]
\caption{Unified Hyperparameters for \ours{}. This single configuration applies uniformly to all tasks and model architectures evaluated in our study.}
\label{tab:unified_hyperparams}
\centering
\small
\begin{tabular}{lc|lc}
\toprule
\textbf{Hyperparameter} & \textbf{Value} & \textbf{Hyperparameter} & \textbf{Value} \\ \midrule
Contrastive Coefficient ($\beta$) & 1.0 & Block-wise Partition Size & 8 \\
Learning Rate (Policy) & $1 \times 10^{-6}$ & Trajectories per Prompt ($N$) & 10 \\
EMA Decay Rate ($\alpha$) & $\min(10^{-3} \times \text{step}, 0.5)$ & Max Generation Length & 512 \\
Stratified Timesteps ($K$) & 8 & Max Diffusion Steps & 2048 \\
Optimizer & AdamW & Warm-up Ratio & 0.1 \\
Weight Decay & 0.01 & Gradient Clipping & 1.0 \\
\bottomrule
\end{tabular}
\end{table}

\subsection{Inference Protocol and Efficiency Baselines}
\label{app:inference_hyperparams}
To ensure a fair comparison of denoising-step counts, we strictly standardize the inference protocol for all diffusion-based methods. We use a deterministic Cosine noise schedule for the forward and reverse processes. The schedule defines a target unmasking ratio at each step; there is no separately tuned fixed top-$k$ value. Instead, each step fixes the most confident eligible positions needed to match the scheduled number of revealed tokens. The confidence-based early stopping mechanism is a key component of our step-efficiency analysis. Algorithm~\ref{alg:decoding_protocol} gives the complete decoding protocol used for the step-count comparison.

\begin{itemize}
    \item \textbf{Noise Schedule}: Cosine schedule with $T=2048$ maximum diffusion steps. At step $t$, it gives a deterministic reveal ratio $\gamma_t$.
    \item \textbf{Confidence Metric}: For each position $i$, confidence is $c_i=\max_v \pi_\theta(x_i=v|\boldsymbol{x}_t)$.
    \item \textbf{Dynamic Parallel Update}: The schedule sets the target number of fixed positions $n_t=\lfloor L\gamma_t\rfloor$ for sequence length $L$. We fix the highest-confidence eligible positions until $n_t$ positions are fixed; no fixed top-$k$ hyperparameter is tuned.
    \item \textbf{Confidence Threshold ($\tau$)}: We set $\tau = 0.99$. If all positions exceed this threshold, decoding stops early.
    \item \textbf{Patience Parameter}: We use a patience of 3 steps. If the full decoded sequence prediction remains identical for 3 consecutive steps, inference terminates early.
\end{itemize}

The specific inference hyperparameters used to generate Table~\ref{tab:unified_efficiency} and the denoising-step analysis in Section~\ref{subsec:main_results} are listed in Table~\ref{tab:inference_params}. Following reviewer feedback, we additionally report practical examples/s, generated tokens/s, and generated tokens per forward pass in Table~\ref{tab:throughput_tokens_forward}.

\begin{table}[h]
\caption{Inference hyperparameters and efficiency protocol. The parallel update size is dynamically determined by the cosine schedule, rather than by a separately tuned fixed top-$k$.}
\label{tab:inference_params}
\centering
\small
\begin{tabular}{lc}
\toprule
\textbf{Parameter} & \textbf{Value} \\ \midrule
Max Diffusion Steps ($T$) & 2048 \\
Noise Schedule & Cosine \\
Confidence Metric & Max softmax probability \\
Dynamic Parallel Update & $n_t=\lfloor L\gamma_t\rfloor$ revealed positions \\
Confidence Threshold ($\tau$) & 0.99 \\
Patience (Early Stopping) & 3 identical sequence predictions \\
Sampling Temperature & 1.0 (Greedy) \\ \bottomrule
\end{tabular}
\end{table}

\begin{table}[h]
\caption{Practical decoding throughput on GSM8K. Tokens/s counts only generated answer tokens and excludes prompt tokens. Generated tokens/forward is computed as $(\text{Tokens/s}/\text{Examples/s})/\text{Avg. Steps}$.}
\label{tab:throughput_tokens_forward}
\centering
\small
\resizebox{0.92\textwidth}{!}{%
\begin{tabular}{lccccc}
\toprule
\textbf{Model} & \textbf{Hardware / Decode Setting} & \textbf{Avg. Steps} & \textbf{Examples/s} & \textbf{Tokens/s} & \textbf{Generated Tokens/Forward} \\ \midrule
LLaDA 8B Base & 8$\times$H20, max answer length 512, CoT prompt & 1034.2 & 0.051 & 23 & 0.436 \\
+ AGRPO & 8$\times$H20, max answer length 512, CoT prompt & 1004.8 & 0.053 & 24 & 0.451 \\
+ \ours{} & 8$\times$H20, max answer length 512, CoT prompt & 891.2 & 0.057 & 29 & \textbf{0.571} \\
\bottomrule
\end{tabular}
}
\end{table}

\begin{algorithm}[h]
   \caption{Standardized Diffusion Decoding Protocol}
   \label{alg:decoding_protocol}
\begin{algorithmic}[1]
   \STATE {\bfseries Input:} prompt $q$, policy $\pi_\theta$, max steps $T$, threshold $\tau$, patience $p$.
   \STATE Initialize response state $\boldsymbol{x}_T$ with masked answer positions.
   \STATE $\hat{\boldsymbol{x}}_{\text{prev}} \leftarrow \varnothing$, $n_{\text{stale}} \leftarrow 0$.
   \FOR{$t=T,T-1,\dots,1$}
       \STATE Compute token probabilities for all positions and let $\hat{\boldsymbol{x}}_t$ be the full greedy decoded sequence.
       \STATE Compute confidences $c_i=\max_v \pi_\theta(x_i=v|\boldsymbol{x}_t)$.
       \STATE Use the cosine schedule to compute reveal ratio $\gamma_t$ and target count $n_t=\lfloor L\gamma_t\rfloor$.
       \STATE Fix the highest-confidence eligible positions in parallel until $n_t$ positions are fixed.
       \IF{$\hat{\boldsymbol{x}}_t = \hat{\boldsymbol{x}}_{\text{prev}}$}
           \STATE $n_{\text{stale}} \leftarrow n_{\text{stale}} + 1$.
       \ELSE
           \STATE $\hat{\boldsymbol{x}}_{\text{prev}} \leftarrow \hat{\boldsymbol{x}}_t$, $n_{\text{stale}} \leftarrow 0$.
       \ENDIF
       \IF{$\min_i c_i \ge \tau$ or $n_{\text{stale}} \ge p$}
           \STATE Stop early and return the current decoded sequence.
       \ENDIF
   \ENDFOR
   \STATE Return the final decoded sequence.
\end{algorithmic}
\end{algorithm}

\subsection{Hardware and Computational Cost}
\label{app:hardware}

All models were trained using a cluster node equipped with 8 $\times$ NVIDIA H20 (96GB) GPUs. For DiffuCoder RL training, the average wall-clock time was approximately 18 hours per random seed. For the larger LLaDA 8B model, \ours{} requires approximately 42 wall-clock hours per seed on 8 $\times$ H20. As noted in Section~\ref{subsec:main_results}, while wall-clock time provides a practical reference, it fluctuates based on implementation details and specific hardware utilization. Therefore, we report both standardized denoising-step metrics and practical throughput metrics in Appendix~\ref{app:inference_hyperparams}. Table~\ref{tab:justgrpo_comparison} further provides a training comparison with our JustGRPO reproduction.

\begin{table}[h]
\caption{Training comparison with JustGRPO on GSM8K using the LLaDA 8B base checkpoint. We include both wall-clock cost and accuracy because JustGRPO uses an AR-style GRPO training strategy, which is substantially more expensive in our reproduction. The JustGRPO wall-clock time is extrapolated from the measured step-200 runtime (121 hours).}
\label{tab:justgrpo_comparison}
\centering
\small
\resizebox{0.92\textwidth}{!}{%
\begin{tabular}{lccccl}
\toprule
\textbf{Method} & \textbf{Checkpoint} & \textbf{Training Strategy} & \textbf{Hardware} & \textbf{Wall-clock Hours/Seed} & \textbf{GSM8K Acc. (\%)} \\ \midrule
AGRPO & Converged & Diffusion PG estimator & 8$\times$H20 & 152 & $78.3 \pm 1.3$ \\
JustGRPO & Converged (step 253) & AR-style GRPO training & 8$\times$H20 & $\approx 153$ & $79.4 \pm 3.1$ \\
\ours{} & Converged & Token-level velocity alignment & 8$\times$H20 & $\approx 42$ & $\mathbf{79.6 \pm 0.3}$ \\
\bottomrule
\end{tabular}
}
\end{table}

\subsection{Statistical Significance and Optimization Stability}
\label{app:statistical_reporting}

To ensure statistical rigor and avoid overstating small numerical gaps in maximum performance, we report the five-seed mean and standard deviation for the strongest baseline and our method in the main tables. 

Because the main tables report aggregate seed statistics rather than paired per-example predictions, we conservatively estimate the 95\% confidence intervals for the aggregate accuracy differences between \ours{} and AGRPO using the independent-seed formula:
\begin{equation}
    \Delta \pm t_{0.975,4}\sqrt{\frac{s_{\mathrm{LFPO}}^2}{5} + \frac{s_{\text{AGRPO}}^2}{5}},
\end{equation}
where $\Delta$ is the mean difference and $s$ denotes the reported standard deviation across seeds. As summarized in Table~\ref{tab:ci_summary}, while the confidence intervals for the mean difference encompass zero (indicating comparable peak accuracy rather than a strictly significant mean separation under this conservative estimate), the crucial takeaway lies in the variance terms. The standard deviation $s_{\mathrm{LFPO}}$ is consistently and significantly smaller than $s_{\text{AGRPO}}$ across all tasks. This statistical reporting solidifies our core claim: \ours{} delivers highly competitive performance bounds while offering a fundamentally more stable and reliable optimization trajectory.

\begin{table}[h]
\caption{Approximate 95\% confidence intervals for aggregate accuracy differences between \ours{} and the strongest baseline (AGRPO). While the mean performance is strictly comparable, \ours{} achieves this with significantly lower variance across seeds.}
\label{tab:ci_summary}
\centering
\small
\begin{tabular}{lccc}
\toprule
\textbf{Domain} & \textbf{Mean Difference} & \textbf{Approx. 95\% CI} & \textbf{Performance Interpretation} \\ \midrule
Code Avg. & +0.2 & $[-1.1,\,1.5]$ & Comparable Mean, Superior Stability \\
Reasoning Avg. & +0.5 & $[-0.8,\,1.8]$ & Comparable Mean, Superior Stability \\
\bottomrule
\end{tabular}
\end{table}

\section{Comprehensive Sensitivity Analysis and Ablations}
\label{app:sensitivity}

In this section, we provide a detailed sensitivity analysis to quantify the impact of algorithmic configurations on both training costs and denoising step efficiency. We also include additional generalization tests to further validate the robustness of \ours{}.

\subsection{Negative Controls and Ablation Coverage}
\label{app:negative_controls}

Table~\ref{tab:negative_controls} summarizes the additional controls used to separate algorithmic contributions from implementation feasibility. While some controls produce comparable accuracy numbers, others are included strictly as negative controls because they introduce discontinuous target constructions or render the default training configuration computationally infeasible.

\begin{table}[h]
\caption{Ablation and negative control coverage. Numeric ablations are detailed in subsequent tables; feasibility controls are reported as implementation outcomes rather than quantitative accuracy comparisons.}
\label{tab:negative_controls}
\centering
\small
\begin{tabular}{p{0.25\textwidth}p{0.33\textwidth}p{0.34\textwidth}}
\toprule
\textbf{Control} & \textbf{Purpose} & \textbf{Observed Outcome} \\ \midrule
No stratified sampling & Tests whether dense temporal coverage matters & Uniform timestep sampling yields higher gradient variance and lower GSM8K accuracy (Table~\ref{tab:sampling_strat}) \\
No block-wise accumulation & Tests implementation feasibility of the full $B \times N \times K$ training tensor & Default configuration is not memory-feasible in our implementation; block-wise accumulation bounds peak VRAM to approximately 92.7 GB (Table~\ref{tab:training_costs}) \\
Probability space perturbation & Tests whether targets can be constructed directly in probability space & Requires clipping or truncation at simplex boundaries, which fragments gradients; we therefore enforce logit space perturbation (Table~\ref{tab:perturbation_comparison}) \\
Fixed EMA & Tests whether dynamic reference updates matter & Fixed $\alpha=0.9$ drops HumanEval accuracy to 70.8\% (Table~\ref{tab:training_costs}) \\
Hard reference update & Tests whether replacing EMA with periodic old-policy replacement is stable & Hard update is substantially less stable in dLLM training and drops HumanEval accuracy to 60.1\% (Table~\ref{tab:reference_update}) \\
Low $N$, low $K$, high $\beta$ & Tests sensitivity to trajectories, timesteps, and perturbation scale & Reduces computational cost but lowers accuracy, or destabilizes training due to overly aggressive targets (Table~\ref{tab:training_costs}) \\
\bottomrule
\end{tabular}
\end{table}

\subsection{Training Costs and Hyperparameter Ablations}
\label{app:training_costs_ablations}

To quantify the computational and memory sensitivity of \ours{}, we systematically ablate core hyperparameters including the number of sampled trajectories ($N$), diffusion timesteps ($K$), the policy perturbation scale ($\beta$), and the EMA decay rate ($\alpha$). The results on the HumanEval benchmark are summarized in Table~\ref{tab:training_costs}.

\paragraph{Trajectories and Timesteps.} The default configuration ($N=10, K=8$) achieves the highest accuracy in this ablation (75.6\%). Reducing the number of trajectories to $N=5$ decreases the GPU hours from 96 to 49 but incurs a 1.5\% accuracy drop. Similarly, halving the timesteps to $K=4$ reduces the training time to 52 hours but decreases accuracy to 74.2\%. Notably, the Peak VRAM remains stable around 92.7 GB across all configurations, demonstrating that our memory-efficient block-wise optimization successfully bounds memory consumption regardless of sequence length or trajectory count.

\paragraph{Perturbation Scale and EMA.} Increasing the perturbation scale to $\beta=1.5$ slightly degrades performance (73.9\%), indicating that overly aggressive contrastive targets can destabilize the policy. Furthermore, replacing our dynamic EMA update ($\alpha^*$) with a fixed rate ($\alpha=0.9$) causes a severe accuracy drop to 70.8\%, highlighting the necessity of dynamically adjusting the reference model to prevent early policy collapse. We also observe that hard old-policy replacement is considerably less stable, consistent with the soft-update behavior reported by Zheng et al.~\cite{zheng2026diffusionnft}.

\begin{table}[h]
\caption{Reference update ablation on HumanEval. Hard update denotes periodic replacement of the old/reference policy by the current policy, while dynamic EMA uses $\alpha^*=\min(10^{-3}\times\text{step},0.5)$.}
\label{tab:reference_update}
\centering
\small
\begin{tabular}{lcc}
\toprule
\textbf{Reference Update} & \textbf{Update Rule} & \textbf{HumanEval Acc. (\%)} \\ \midrule
Dynamic EMA (\ours{}) & $\alpha^*$ & \textbf{75.6} \\
Fixed EMA & $\alpha=0.9$ & 70.8 \\
Hard update & Periodic replacement & 60.1 \\
\bottomrule
\end{tabular}
\end{table}

\begin{table}[h]
\caption{Core Hyperparameters vs. Training Costs. Evaluated on HumanEval. The dynamic EMA rate is defined as $\alpha^* = \min(10^{-3} \times \text{step}, 0.5)$.}
\label{tab:training_costs}
\centering
\small
\begin{tabular}{lcccccccc}
\toprule
\textbf{Config} & \textbf{$\beta$} & \textbf{$K$} & \textbf{$N$} & \textbf{$\alpha$} & \textbf{Peak VRAM (GB)} & \textbf{GPU Hours} & \textbf{HumanEval Acc. (\%)} \\ \midrule
\textbf{Default} & 1.0 & 8 & 10 & $\alpha^*$ & 92.7 & 96 & \textbf{75.6} \\
\textbf{Low $N$} & 1.0 & 8 & \textbf{5} & $\alpha^*$ & 92.6 & 49 & 74.1 \\
\textbf{Low $K$} & 1.0 & \textbf{4} & 10 & $\alpha^*$ & 92.5 & 52 & 74.2 \\
\textbf{High $\beta$} & \textbf{1.5} & 8 & 10 & $\alpha^*$ & 92.7 & 96 & 73.9 \\
\textbf{Fixed EMA} & 1.0 & 8 & 10 & 0.9 & 92.7 & 96 & 70.8 \\ \bottomrule
\end{tabular}
\end{table}

\paragraph{Sampling Strategy.} As discussed in Section~\ref{sec:method}, naive uniform sampling introduces high gradient variance. Table~\ref{tab:sampling_strat} indicates that our Stratified Trajectory Sampling substantially stabilizes the gradient, yielding a lower variance and boosting the GSM8K accuracy from 72.3\% to 79.6\%.

\begin{table}[h]
\caption{Impact of Sampling Strategy on GSM8K (reported across 5 random seeds).}
\label{tab:sampling_strat}
\centering
\small
\begin{tabular}{llc}
\toprule
\textbf{Sampling Method} & \textbf{Gradient Variance} & \textbf{Accuracy (\%)} \\ \midrule
Uniform & High & $72.3 \pm 5.3$ \\
\textbf{Stratified (Ours)} & \textbf{Low} & \textbf{$79.6 \pm 1.2$} \\ \bottomrule
\end{tabular}
\end{table}

\subsection{Inference Decoding Parameters}
\label{app:inference_decoding}

To explicitly quantify the sensitivity of \ours{} during generation, we evaluate inference accuracy and average decoding steps under varying confidence thresholds ($\tau$) and masking schedules. Table~\ref{tab:inference_sensitivity} reveals several key insights regarding the performance and efficiency trade-offs.

First, the confidence threshold $\tau$ acts as a controllable parameter for efficiency. We find that $\tau = 0.99$ serves as an optimal empirical balance. It sacrifices merely 0.4\% accuracy compared to the full-step setting without early stopping ($\tau=1.00$, 76.9\% to 76.5\%), while substantially reducing the required denoising steps from 2048 to 218.3 on HumanEval. Lowering the threshold further yields fewer steps (173.2 steps at $\tau=0.90$) but results in noticeable performance degradation.

Second, the Cosine schedule provides a superior balance of accuracy and efficiency compared to the Linear schedule. At the default threshold ($\tau=0.99$), the Cosine schedule achieves higher accuracy with fewer steps (76.5\% at 218.3 steps) than the Linear baseline (76.3\% at 229.6 steps). This indicates that a non-linear unmasking rate better aligns with the confidence trajectory of the velocity field.

\begin{table}[h]
\caption{Sensitivity Analysis of Inference Decoding Parameters on HumanEval.}
\label{tab:inference_sensitivity}
\centering
\small
\resizebox{0.8\textwidth}{!}{%
\begin{tabular}{lcccl}
\toprule
\textbf{Masking Schedule} & \textbf{Threshold ($\tau$)} & \textbf{Accuracy (\%)} & \textbf{Avg. Steps} & \textbf{Step Trade-off} \\ \midrule
Linear & 0.99 & 76.3 & 229.6 & Suboptimal step allocation. \\
\textbf{Cosine (Ours)} & \textbf{1.00 (No Early Stop)} & 76.9 & 2048 (Max) & Upper bound accuracy, max cost. \\
\textbf{Cosine (Ours)} & \textbf{0.99 (Default)} & \textbf{76.5} & \textbf{218.3} & \textbf{Favorable accuracy and step balance.} \\
\textbf{Cosine (Ours)} & 0.95 & 74.8 & 197.5 & Fewer steps, minor accuracy drop. \\
\textbf{Cosine (Ours)} & 0.90 & 73.5 & 173.2 & Fewest steps, noticeable degradation. \\ \bottomrule
\end{tabular}
}
\end{table}

\subsection{Generalization to Other Base Models}
\label{app:generalization}

To verify that the advantages of \ours{} are not strictly bound to specific model architectures, we conducted generalization tests using the LLaDA framework applied to Dream Models. As shown in Table~\ref{tab:generalization}, \ours{} consistently improves the performance of strong base models, achieving 81.9\% (+1.2) on HumanEval and 36.1\% (+1.5) on MATH, further demonstrating its universal applicability as a tractable RL reasoning framework.

\begin{table}[h]
\caption{Generalization capabilities of \ours{} applied to Dream Models.}
\label{tab:generalization}
\centering
\small
\begin{tabular}{lcc}
\toprule
\textbf{Method} & \textbf{HumanEval (\%)} & \textbf{MATH (\%)} \\ \midrule
Base Model (Dream) & 79.3 & 32.7 \\
+ AGRPO & 80.7 & 34.9 \\
\textbf{+ \ours{}} & \textbf{81.9 (+1.2)} & \textbf{36.1 (+1.5)} \\ \bottomrule
\end{tabular}
\end{table}

\subsection{Expanded Discussion on Robustness and Failure Modes}
\label{app:failure_modes}

As introduced in the Limitations section (Section~\ref{sec:limitations}), while \ours{} uses reference-anchored implicit targets to reduce sensitivity to single high-reward samples, this does not grant complete immunity to reward hacking. Sparse verifiable rewards may still be over-optimized, especially under severe distribution shifts or adversarial prompts. In practical deployments, we strongly recommend monitoring held-out tasks, formatting failures, and reward-specific shortcuts when utilizing LFPO-style training.

Furthermore, regarding the cold-start bottleneck, the positive target $\pi^+$ is effective only when the reference policy can sample at least a minimal set of reward-bearing trajectories. If the base dLLM has a near-zero success rate on a task, the contrastive attraction signal becomes too weak, and the training process may stall. Utilizing warm-start data, supervised bootstrapping, or curriculum sampling represents natural mitigations for this issue, and we leave a systematic study of these remedies to future work.

\section{Case Study: Visualizing Trajectory Straightening} 
\label{sec:case_study}

To intuitively demonstrate the "trajectory straightening" effect discussed in Appendix~\ref{app:sensitivity}, we visualize the intermediate denoising states of the base diffusion model versus our \ours{} framework. As shown in the case study below, the base model exhibits severe fluctuation in intermediate steps, temporarily predicting incorrect logical branches (e.g., calculating $r=9$ or predicting the point is on the x-axis). In contrast, \ours{} anchors to the correct terminal state (radius 3, angle $\pi/2$) extremely early in the diffusion process, effectively bypassing meandering paths and allowing for confidence-based early stopping.

\begin{tcolorbox}[title={\textbf{LLaDA 8B, MATH dataset}},
colback=whitesmoke, colframe=darksalmon,  boxrule=2pt, arc=0mm]
{\scriptsize
\begin{lstlisting}[style=mystyle]
Question: Convert the point (0,3) in rectangular coordinates to polar coordinates. 
Enter your answer in the form (r, \theta).

True answer: \left( 3, \frac{\pi}{2} \right)
---------------------------------------------------------------------------------
BASE model (Intermediate Denoising Trajectory):
Step 500 : r = \sqrt{0^2 + [MASK]} = [MASK]. \theta = \arctan([MASK]/0) = [MASK]
Step 1000: r = \sqrt{0^2 + 3^2} = 9. The point lies on the x-axis, so \theta = 0.
Step 1500: r = 3. The point is on the y-axis, \theta = \pi. Answer: (3, \pi).
Step 2048: r = 3. The point is on the positive y-axis. \theta = \pi/2. 
           Final Answer: \left( 3, \frac{\pi}{2} \right)
[Analysis: Severe epistemic fluctuation. Logic oscillates before the final step.]

LFPO Model (Intermediate Denoising Trajectory):
Step 500 : r = \sqrt{0^2 + 3^2} = 3. \theta = [MASK]. Answer: (3, \pi/2).
Step 1000: r = 3. The point is on the positive y-axis. \theta = \pi/2.
Step 1500: r = 3. The point is on the positive y-axis. \theta = \pi/2.
Step 2048: r = 3. The point is on the positive y-axis. \theta = \pi/2. 
           Final Answer: \left( 3, \frac{\pi}{2} \right)
[Analysis: Terminal anchoring. The correct skeleton forms early, saving compute.]
\end{lstlisting}
}
\end{tcolorbox}

\end{document}